\pdfoutput=1

\documentclass[11pt]{article}

\usepackage{EACL2024}

\usepackage{times}
\usepackage{latexsym}

\usepackage{array}
\usepackage{color}
\usepackage{bm}
\usepackage{bbm}
\usepackage{enumitem}
\usepackage{booktabs}
\usepackage{amssymb}
\usepackage{amsfonts}
\usepackage{amsmath}
\usepackage{multirow}
\usepackage{graphicx}
\usepackage{amsfonts}
\usepackage[linesnumbered,ruled,vlined]{algorithm2e}
\usepackage{colortbl}
\usepackage{CJKutf8}
\usepackage{url}
\usepackage{here}
\usepackage{subcaption}
\usepackage{makecell}
\usepackage{mathrsfs}
\usepackage{titlesec}
\usepackage{lipsum}
\usepackage{booktabs}
  
\newcommand*{\Ja}[1]{%
  \begin{CJK}{UTF8}{ipxm}#1\end{CJK}}

\DeclareMathOperator*{\stat}{stat}
\DeclareMathOperator*{\mean}{mean}

\usepackage[T1]{fontenc}

\usepackage[utf8]{inputenc}

\usepackage{microtype}

\usepackage{inconsolata}

\title{Generating Diverse Translation with Perturbed $k$NN-MT}

\author{
  Yuto Nishida$^{1}$ $\;\;\;$ 
  Makoto Morishita$^{2}$ $\;\;\;$ 
  Hidetaka Kamigaito$^{1}$  $\;\;\;$ 
  Taro Watanabe$^{1}$ \\
  $^1$Nara Institute of Science and Technology $\;$ \\
  $^2$NTT Communication Science Laboratories, NTT Corporation$\;$ \\
  \texttt{\{nishida.yuto.nu8, kamigaito.h, taro\}@is.naist.jp} \\
  \texttt{makoto.morishita@ntt.com}
}

\begin{document}
\maketitle
\begin{abstract}
Generating multiple translation candidates would enable users to choose the one that satisfies their needs.
Although there has been work on diversified generation, there exists room for improving the diversity mainly because the previous methods do not address the overcorrection problem---the model underestimates a prediction that is largely different from the training data, even if that prediction is likely.
This paper proposes methods that generate more diverse translations by introducing perturbed $k$-nearest neighbor machine translation ($k$NN-MT).
Our methods expand the search space of $k$NN-MT and help incorporate diverse words into candidates by addressing the overcorrection problem.
Our experiments show that the proposed methods drastically improve candidate diversity and control the degree of diversity by tuning the perturbation's magnitude.
\end{abstract}

\section{Introduction}
In natural language, there are multiple lexically distinct translations given an input sentence.
Therefore, machine translation systems should offer multiple translation candidates to users so that the final choice should be made by them considering their demands, e.g., styles or domains.
However, standard neural machine translation (NMT) models suffer from a low diversity problem in which the generated translation candidates are almost identical.
One reason lies in beam search, which is a standard inference algorithm, where the search space is expanded in a left-to-right fashion while keeping only the top-$N$ candidates in every decoding step and just preserving slightly different translations~\citep{gimpel-etal-2013-systematic,Vijayakumar_Cogswell_Selvaraju_Sun_Lee_Crandall_Batra_2018,freitag-al-onaizan-2017-beam}.
The other reason is the overcorrection problem~\citep{zhang-etal-2019-bridging}, which is caused by a model trained with cross-entropy loss that underestimates a prediction that is largely different from the training data, even if it is likely.
This phenomenon discourages the model from generating synonymous expressions and leans toward gold standards, reducing the diversity in the candidates.

To encourage the model to generate more diverse candidates, \citet{Vijayakumar_Cogswell_Selvaraju_Sun_Lee_Crandall_Batra_2018}, \citet{Holtzman2020The}, and \citet{freitag-al-onaizan-2017-beam} proposed variants of beam search algorithms in which diverse candidates are retained in the search space.
However, their methods do not directly address the overcorrection problem, limiting their effect in generating diverse translations.

 \begin{figure*}[htbp]
    \centering
    \includegraphics[width=1.0\linewidth]{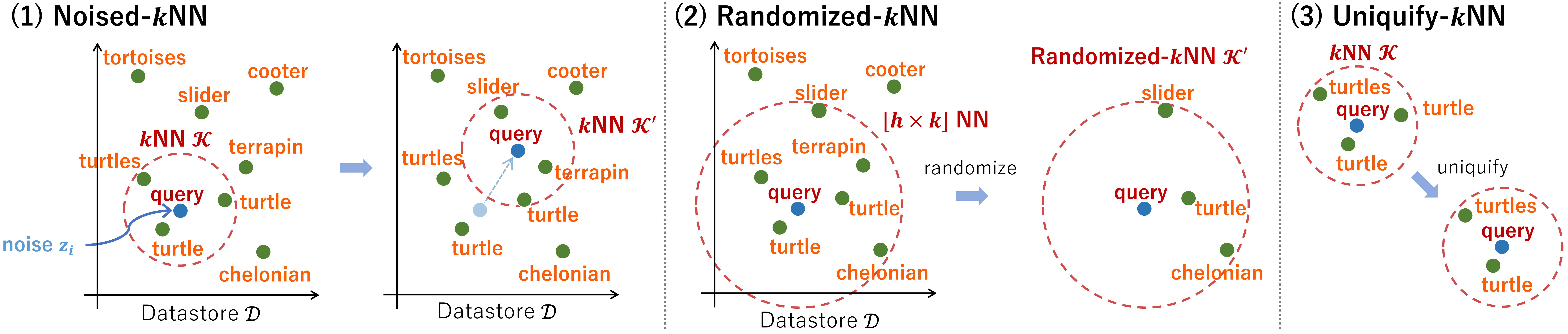}
    \caption{Overview of our proposed method: See \S \ref{sec:proposed_method} for details. Green points represent target tokens in datastore. Blue points represent query vectors, and surrounding circles denote retrieved neighbors. (1) Noised-$k$NN adds a noise vector to the query, changing the retrieved tokens. (2) Randomized-$k$NN initially retrieves more neighbors and randomly selects $k$-neighbors. (3) Uniquify-$k$NN only considers unique target tokens from retrieved neighbors. In this figure, number of neighbors $k$ per query is set to $3$, and hyperparameter $h$ of (2)~Randomized-$k$NN is set to $2.0$.}
    \label{fig:proposed}
\end{figure*}

To alleviate this issue, we propose $k$NN diversified decoding that combines diversified beam search and $k$-nearest neighbor machine translation~($k$NN-MT; \citealp{khandelwal2021nearest}), which addresses the overcorrection problem by retrieving alternative target tokens from the training data during decoding~\cite{yang-etal-2022-nearest}.
To further diversify the search space, we also propose two methods, i.e., stochastic and deterministic methods.
The stochastic method expands the search space by perturbation so that the model can generate more likely tokens that are less focused.
We proposed two types of perturbations, \emph{noised-$k$NN} (Figure~\ref{fig:proposed}~(1)), which adds a noise vector to the query of the $k$NN search, and \emph{randomized-$k$NN} (Figure~\ref{fig:proposed}~(2)), which arbitrarily selects $k$ neighbors from a more extensive search space.
The deterministic method, \emph{uniquify-$k$NN} (Figure~\ref{fig:proposed}~(3)), removes duplicates from the retrieved $k$NN tokens so that no token can be dominant and thus more diverse candidates remain.

Our experiments showed that our proposed methods alleviate the overcorrection problem that leads to the generation of more diverse candidates, and maintain fluency and oracle translation quality in multiple domains and language pairs.
We also show that the degree of diversity can be controlled by changing the perturbation's magnitude, which benefits end-applications, e.g., human post-editing.

\section{Related Work}
\subsection{Diverse Text Generation}
\label{sec:related_diversity}
Given the importance of generating diverse translations, many of the proposed search algorithm variations can be categorized into either deterministic or stochastic types.

For the former method, \citet{Vijayakumar_Cogswell_Selvaraju_Sun_Lee_Crandall_Batra_2018} proposed diverse beam search (DBS) in which beams are divided into several groups, and a modified score function penalizes the overlapped tokens among the groups.
\citet{freitag-al-onaizan-2017-beam} proposed a method that determines the maximum number of candidates that share the same partial hypothesis.

As for the latter approach, top-$k$ sampling~\citep{fan-etal-2018-hierarchical} randomly samples the output tokens from the top $k$-tokens with the highest likelihood at each time step.
Similarly, nucleus sampling~\citep{Holtzman2020The} randomly samples from the smallest subset of candidates whose total likelihood exceeds $p$ at each time step.
Noisy parallel approximate decoding~\citep{cho2016noisy} explores multiple modes by injecting noise into the model's hidden states.
\citet{wu-etal-2020-generating} proposed a method that samples different models derived by applying concrete dropout.
MixDiversity~\cite{li-etal-2021-mixup-decoding} leverages the hidden representations of the randomly sampled sentence pairs from the subset of the training corpus.

Although all of the above methods diversify the output text, they do not explicitly address the overcorrection problem, which is the root cause of the limited diversity~(\S \ref{sec:related_overcorrection}).

\subsection{\texorpdfstring{$k$NN-MT}{}}
\label{sec:related_knnmt}
\citet{khandelwal2021nearest} proposed $k$-nearest neighbor machine translation ($k$NN-MT), which uses $k$NN search for retrieving similar examples during inference by reflecting the retrieval results in the score function.
The translation quality is improved by allowing the model to directly access large-scale cached translation examples.
$k$NN-MT consists of two steps, datastore creation and generation.

\paragraph{Datastore creation}
Before inferences with $k$NN-MT, we need to create a datastore, i.e., key-value pairs of high-dimensional representations and tokens.
We feed all the training data into the NMT model and save each target token as a value and its decoder hidden state vector as a key representation.
Formally, let $f(\boldsymbol{x}, \boldsymbol{y}_{<i})$ be the hidden state vector at time step $i$ for source sentence $\boldsymbol{x} \in \mathcal{S}$ and target sentence $\boldsymbol{y} \in \mathcal{T}$ of training data $(\mathcal{S}, \mathcal{T})$, and then datastore $\mathcal{D}$ can be represented:
\begin{multline}
\label{eq:datastore}
    \mathcal{D} = 
    \{(f(\boldsymbol{x}, \boldsymbol{y}_{<i}),~ y_i), ~\forall{y_i} \in \boldsymbol{y} \\~|~ (\boldsymbol{x}, \boldsymbol{y}) \in (\mathcal{S}, \mathcal{T}) \}.
\end{multline}

\paragraph{Generation}
To generate a sentence from a given input sentence $\boldsymbol{x}$, we extract $k$-nearest neighbors ${\mathcal{K} \subset \mathcal{D}}$ from the datastore using decoder hidden state $\boldsymbol{q}_{i}$ at time step $i$ as a query corresponding to output token $y_{i}$.
$k$-nearest neighbor probability $p_{\text{$k$NN}}$ is calculated from the distances between query $\boldsymbol{q}_{i}$ and the $k$-nearest neighbors:
\begin{align}
\label{eq:kNN_probability}
    & p_{\text{$k$NN}}(y_i|\boldsymbol{x}, \boldsymbol{y}_{<i}) \propto \nonumber \\
    & \sum_{(\boldsymbol{k}_{j},v_j) \in \mathcal{K}}{\mathbbm{1}_{y_i=v_j}\exp \left(\frac{-\mathrm{dist}(\boldsymbol{k}_{j},\boldsymbol{q}_{i})}{\tau}\right )},
\end{align}
where $\mathrm{dist}(\cdot,\cdot)$ is a distance function and $\tau$ is a softmax temperature parameter.
The word probability of $y_{i}$ is calculated by the linear interpolation of $k$NN probability $p_{\text{$k$NN}}$ and output probability $p_{\text{MT}}$ of the NMT model:
\begin{align}
\label{eq:interpolation}
    & p(y_i|\boldsymbol{x}, \boldsymbol{y}_{<i}) = \nonumber \\
    & \lambda p_{\text{$k$NN}}(y_i|\boldsymbol{x}, \boldsymbol{y}_{<i})
    + (1 - \lambda)p_{\text{MT}}(y_i|\boldsymbol{x}, \boldsymbol{y}_{<i}),
\end{align}
where $\lambda$ is a hyperparameter that determines the weight of the $k$NN probability.

$k$NN-MT substantially improves the translation performance without additional model training, and several variants have been proposed.
\citet{jiang-etal-2021-learning,zheng-etal-2021-adaptive,jiang-etal-2022-towards} further improved the translation performance by dynamically changing the number of neighbors and the interpolation weight.
\citet{wang-etal-2022-efficient, meng-etal-2022-fast, deguchi-etal-2023-subset} proposed methods for faster inference by reducing search space.
However, no research uses $k$NN-MT for improving generation diversity.

\subsection{Overcorrection}
\label{sec:related_overcorrection}
The standard NMT models trained with cross-entropy loss suffer from the overcorrection problem~\citep{zhang-etal-2019-bridging} in which the model underestimates a prediction that is largely different from the training data, even if it is likely.
We hypothesize that this problem decreases the diversity of candidates due to the low probabilities for alternative tokens assigned by the underlying model.

\citet{zhang-etal-2019-bridging} alleviated overcorrection by mitigating the discrepancy between training and inference.
\citet{yang-etal-2022-nearest} argued that the $k$NN-MT's improvement is derived from alleviating the overcorrection problem by a $k$NN search.
However, the relationship between overcorrection and generation diversity remains unclear.
In this study, we propose expanding the search space of kNN-MT to alleviate overcorrection.
We also conduct a quantitative analysis of overcorrection and diversity~(\S\ref{sec:overcorrection_analysis}).

\section{\texorpdfstring{$k$NN Diversified Decoding}{}}
\label{sec:proposed_method}

We propose to employ $k$NN-MT to alleviate the overcorrection problem and thus encourage diverse generation model-wise.
It is further combined with diversified decoding together with our proposed stochastic and deterministic methods for the more controlled expansion of the search space in $k$NN-MT.\footnote{We combine these methods by calculating word probability with vanilla or perturbed $k$NN-MT and generate candidates by using diversified decoding methods as a search strategy.}

In $k$NN-MT, $k$NN search is expected to improve the output probability of alternative tokens that are not normally included in the top-$N$ of the output probability.
Furthermore, the search space is extensively explored using a diversified decoding method to generate diverse and likely translation candidates.
Although $k$NN search is limited by $k$, more space is explored by stochastically expanding it by adding perturbations from noising (\S \ref{sec:noise}) and randomizing (\S \ref{sec:randomize}).
In addition, deterministically considering only unique tokens in neighbors further allows the model to explore alternative candidates~(\S \ref{sec:uniquify}).

\subsection{\texorpdfstring{Noised-$k$NN}{}}
\label{sec:noise}
As a simple way to perturb the $k$NN distribution, we propose noised-$k$NN, a method that adds a noise vector to the query for a $k$NN search (Figure~\ref{fig:proposed}~(1)).
This method diversifies the candidates by stochastically extending the range of the $k$NN search. 
In this method, we perform a $k$NN search with query ${\boldsymbol{q}_{i}+\boldsymbol{z}_{i}}$, where $\boldsymbol{q}_{i}$ is the hidden decoder states, and $\boldsymbol{z}_{i}$ is the noise vector for output token $y_{i}$ to obtain $k$-nearest neighbors $\mathcal{K}'$.
We then compute the $k$NN probability from $\mathcal{K}'$ in Eq.~\ref{eq:kNN_probability}.
Noise vector $\boldsymbol{z}_{i}$ is generated independently at each time and for each beam as the white Gaussian noise of norm $|a|$ where $a \sim \mathcal{N}(m,s^2)$ with mean $m$ and variance $s^2$.
We propose the following two methods to set $m$ and $s$.

\paragraph{Static noise}
We introduce static noise by setting $m=h_{m}, s=h_{s}$ using hyperparameters $h_{m},h_{s}$.
Hyperparameters $h_{m}$ and $h_{s}$ should be set to appropriate values based on the statistics of the datastore.
In this study, we computed the mean and variance of the distance to the nearest neighbors on the validation data in advance.

\paragraph{Adaptive noise}
As an alternative to static noise, we introduce adaptive noise in which the magnitude of the noises is computed on the fly for each query at each decoding step.
Specifically, a usual $k$NN search is performed to obtain maximum $d_{\text{max}}$ and standard deviation $d_{\text{std}}$ of the distances to the neighbors.
Then an actual noisy $k$NN search is performed by setting $m=h'_{m} \times d_{\text{max}}$ and $s=h'_{s} \times d_{\text{std}}$ using hyperparameters $h'_{m},h'_{s}$.
This method's benefit is that the magnitude of the noise is determined on the fly and eliminates the need for the prior computation of the datastore distributions at the cost of an additional $k$NN search at each decoding step.

\subsection{\texorpdfstring{Randomized-$k$NN}{}}
\label{sec:randomize}
Randomized-$k$NN, as described in Figure~\ref{fig:proposed}~(2), stochastically samples a portion of the expanded neighbors to alleviate the drawbacks of two noising approaches~(\S \ref{sec:noise}) that demand prior computation of parameters $m$ and $s$.
$\lfloor h \times k \rfloor$ neighbors are retrieved where $h$ is a hyperparameter satisfying $h>1$, and randomized $k$-nearest neighbors $\mathcal{K}'$ are obtained by uniformly randomly sampling $k$ from the $\lfloor h \times k \rfloor$ neighbors.
This method is expected to diversify the candidates because it includes more neighbors in the search space.
We do not need to collect any statistics of the distribution of the distances from the query to the $k$-nearest neighbors in advance because we do not perturb the query itself.
In addition, since this method requires only one $k$NN search at each time step, it is identical to the vanilla $k$NN-MT.

\subsection{\texorpdfstring{Uniquify-$k$NN}{}}
\label{sec:uniquify}
The perturbations in \S \ref{sec:noise} and \S \ref{sec:randomize} may have a limited effect on increasing diversity when duplicated tokens are retrieved from the nearest neighbors.
We alleviate this problem by introducing uniquify-$k$NN in which duplicated tokens are explicitly removed from the neighbors (Figure~\ref{fig:proposed} (3)). 

Since the datastore accumulates all the tokens on target-side of the training data, the $k$-nearest neighbors retrieved from the datastore can contain duplicated tokens.
As seen in Eq.~\ref{eq:kNN_probability}, their distance scores are accumulated for duplicated tokens, creating a spuriously dominant probability mass in the neighbor distribution.
Biased probabilities can negatively impact diversity.
Since a larger datastore implies more potential for overlapped tokens, it would further degrade the diversity.

To address this issue, after retrieving the $k$-nearest neighbors, we propose uniquify-$k$NN, a method that eliminates the duplicate tokens from the neighbors, leaving only unique tokens that are closest to the query.
Our new method is formally defined as follows:
\begin{align}
\label{eq:uniquify-kNN}
    & p_{\text{$k$NN}}(y_i|\boldsymbol{x}, \boldsymbol{y}_{<i}) \propto \nonumber \\ 
    & \max_{(\boldsymbol{k}_{j},v_j) \in \mathcal{K}}\mathbbm{1}_{y_i=v_j} \exp \left(\frac{-\mathrm{dist}(\boldsymbol{k}_{j},\boldsymbol{q}_{i})}{\tau}\right ).
\end{align}
This operation prevents the $k$NN probability from becoming peaky and decreasing in diversity.

\section{Experiments}
\label{sec:experiments}
We experimentally confirmed whether our method can generate diverse translation candidates.

\subsection{Experimental Settings}
\subsubsection{Dataset}
The experiments are divided into a domain adaptation setting and a general-domain setting. 
In the domain adaptation setting, we used German-English~(De-En) and Japanese-English (Ja-En) language pairs. 
For De-En, we used five domain data~\cite{koehn-knowles-2017-six,aharoni-goldberg-2020-unsupervised}: Koran, IT, Medical, Law, and Subtitles.
For Ja-En, we used four domain data: the Asian Scientific Paper Excerpt Corpus~(ASPEC; \citealp{nakazawa-etal-2016-aspec}), the Kyoto Free Translation Task~(KFTT; \citealp{neubig11kftt}), TED talks~\cite{cettolo-etal-2012-wit3}, and the Business Scene Dialogue corpus~(BSD; \citealp{rikters-etal-2019-designing}).
We used the designated test set for each domain.

In the general-domain setting, we used three language pairs: WMT'19 news task data~\cite{barrault-etal-2019-findings} for German-English (De-En) and WMT'22 general task data~\cite{kocmi-EtAl:2022:WMT} for Japanese-English (Ja-En) and Ukrainian-Czech~(Uk-Cs).
For the general-domain test set, we used newstest2019 for De-En and generaltest2022 for Ja-En and Uk-Cs.
The statistics of the dataset for both settings are in Appendix~\ref{appendix:detailed_dataset}.

\subsubsection{Models}
\paragraph{Baseline}
All our experiments were carried out with Transformer models~\cite{vaswani17transformer}.
In the domain adaptation and general domain for De-En, we used the WMT'19 De-En pre-trained model~\cite{ng-etal-2019-facebook} available for the \texttt{fairseq} toolkit~\cite{ott-etal-2019-fairseq}.
In the domain adaptation for Ja-En, we used the Transformer Big model trained on JParaCrawl v3.0~\cite{morishita-etal-2022-jparacrawl} as a base model.\footnote{We did not use WMT'22 data for the domain adaptation settings for fair comparisons since it includes KFTT, which is one target domain.}
In the general domain for Ja-En and Uk-Cs, we used Transformer Big models trained on WMT'22 data as a base model for each language pair.
These models were used for datastore creation and as baseline models.
In all the experiments, the beam size was set to 20.

\paragraph{$k$NN-MT}
We used \texttt{FAISS}~\cite{johnson2019billion} for datastore creation and $k$NN search.
The detailed settings are described in Appendix~\ref{appendix:detailed_models}.

\paragraph{Diversified decoding}
We used DBS and nucleus sampling (Nucleus) as the diversified decoding method; the number of DBS groups was set to 20, the diversity strength was set to 0.5, and hyperparameter $p$ of Nucleus was tuned with the validation data.
For our proposed methods, we combined them with DBS and Nucleus.\footnote{From preliminary experiments, we describe the uniquify-$k$NN results in the general-domain setting.}
The hyperparameters of the proposed methods were tuned with the validation data.
The detailed settings are in Appendix~\ref{appendix:detailed_models}.

\subsubsection{Evaluation Metrics}
We used the following metrics to confirm how correctly our model translates and how diverse its candidates are.\footnote{The detailed settings are described in Appendix~\ref{appendix:detailed_metrics}. We also used COMET and BERTScore, but since these scores tend to be similar to BLEU, we show the details and results for these metrics in Appendix~\ref{appendix:detailed_results}.}

\paragraph{BLEU@N}
is a variant of corpus-wise BLEU~\cite{papineni-etal-2002-bleu} computed by the largest sentence-level BLEU score~\cite{chen-cherry-2014-systematic} for each $N$-best candidate, also known as oracle BLEU.
It corresponds to the upper bound of performance through $N$-best reranking.
We report BLEU@1 and BLEU@20 in our experiment.
Note that BLEU@1 is a standard BLEU.

\paragraph{MergedBLEU@N}
is a variant of BLEU@N computed on the merged outputs from two systems.
We employ MergedBLEU@40, which merges 20 candidates from the baseline and a diversified method.
The higher MergedBLEU@40 than BLEU@20 of the baseline implies that the diversified method helps generate the better translations.

\paragraph{Diversity}
The BLEU-based discrepancy metric~(DP; \citealp{shu-etal-2019-generating}) is a measure of the diversity.
DP captures how many unique $n$-grams are included in each candidate sentence, where a higher DP indicates the candidates are diverse.\footnote{As an additional diversity metric, we discuss the number of differences in $n$-gram type in \S\ref{sec:ngram_types}.}

\paragraph{Diversity and translation quality}
The diversity enhancement per quality (DEQ; \citealp{Sun_Huang_Wei_Dai_Chen_2020}) measures the quality-diversity trade-off.
We adapt the DEQ for our experimental settings by using $k$NN-MT as our base:
\begin{align}
    \mathrm{DEQ} = -\frac{\mathrm{DP_{base}} - \mathrm{DP_{sys}}}{\mathrm{RefBLEU_{base}} - \mathrm{RefBLEU_{sys}}}
\end{align}
where $\mathrm{DP_{sys}}$ and $\mathrm{DP_{base}}$ are DP of the evaluated system and $k$NN-MT, respectively, and $\mathrm{RefBLEU_{sys}}$ and $\mathrm{RefBLEU_{base}}$ refer to reference BLEU (RefBLEU; \citealp{Sun_Huang_Wei_Dai_Chen_2020}), the average corpus-wise BLEU across all translation candidates, of the evaluated system and $k$NN-MT, respectively.
The DEQ will be higher if the evaluated system achieves a better quality-diversity trade-off.

\paragraph{Fluency}
The pseudo-log-likelihood score (PLL; \citealp{salazar-etal-2020-masked}) is a metric of fluency using the MLM model.\footnote{We used a multilingual BERT~\cite{devlin-etal-2019-bert} as the MLM model.}
We defined a variant of the PLL for the entire output translations, named SPLL, using statistical function $\stat$:
\begin{align}
\label{eq:MaxPLL}
    \text{SPLL}(\mathbb{W})=\frac{1}{|\mathbb{W}|}\sum_{\mathbf{B} \in \mathbb{W}} \stat_{\hat{\boldsymbol{y}} \in \mathbf{B}}\left( \frac{1}{|\hat{\boldsymbol{y}}|}\text{PLL}(\hat{\boldsymbol{y}}) \right),
\end{align}
where $\mathbb{W} = \{ \mathbf{B}_{1}, \hdots, \mathbf{B}_{M} \}$ is system output, $\mathbf{B}_{k} = \{ \hat{\boldsymbol{y}}^{1}_{k}, \hdots, \hat{\boldsymbol{y}}^{N}_{k} \}$ is the set of $N$-best hypotheses for a source sentence $\boldsymbol{x}_{k} \in \mathcal{X}$, and $\mathcal{X}$ is a test set with $M$ sentences.

In the experiment, to investigate the variances in fluency, we use MaxPLL, MinPLL, and MeanPLL, which use $\max$, $\min$, and $\mean$ functions for the $\stat$ of SPLL.
We also compute the reference's MeanPLL to check how practical the translations' fluency are.
If the generated texts are not as fluent as the reference, the MeanPLL will be lower than the reference.

\subsection{Experimental Results}
\label{sec:results}
\begin{table*}[htbp]
  \centering
  \small
  \begin{tabular}{@{}lcccccc|ccc@{}}
    \toprule
    & Diversity & \multicolumn{4}{c}{Translation Quality (BLEU $\uparrow$ )} & Both & \multicolumn{3}{c}{Fluency (PLL $\uparrow$ )} \\
    Method & DP $\uparrow$ & @1 & @20 & Merged@40 & Ref & DEQ $\uparrow$ & Max & Min & Mean \\
    \cmidrule[.05em](r){1-1} \cmidrule[.05em](lr){2-2} \cmidrule[.05em](lr){3-6} \cmidrule[.05em](lr){7-7} \cmidrule[.05em](l){8-10}
    Reference & - & - & - & - & - & - & - & - & $ -3.35 $ \\
    \cmidrule[.03em](r){1-1} \cmidrule[.03em](lr){2-2} \cmidrule[.03em](lr){3-6} \cmidrule[.03em](lr){7-7} \cmidrule[.03em](l){8-10}
    Baseline & $ 31.4 $ & $ 34.1 $ & $ 42.6 $ & $ 42.6 $ & $ 30.9 $ & $-0.12$ & $ -2.26 $ & $ -4.55 $ & $ \mathbf{-3.28} $ \\
    DBS & $ 35.9 $ & $ 33.6 $ & $ 40.0 $ & $ 43.8 $ & $ 30.3 $ & \phantom{$-$}$ 0.44 $ & $ -2.23 $ & $ -4.63 $ & $ \mathbf{-3.28} $ \\
    Nucleus & $ 48.0 $ & $ 33.4 $ & $ 42.1 $ & $ 44.6 $ & $ 30.0 $ & \phantom{$-$}$ 1.88 $ & $ -2.31 $ & $ \mathbf{-4.42} $ & $ -3.29 $ \\
    $k$NN-MT & $ 32.3 $ & $ \mathbf{43.2} $ & $ \mathbf{51.8} $ & $ \mathbf{53.5} $ & $ \mathbf{38.4} $ & - & $ -2.23 $ & $ -4.74 $ & $ -3.32 $ \\
    \cmidrule[.03em](r){1-1} \cmidrule[.03em](lr){2-2} \cmidrule[.03em](lr){3-6} \cmidrule[.03em](lr){7-7} \cmidrule[.03em](l){8-10}
    DBS+$k$NN-MT & $ 42.0 $ & $ 42.0 $ & $ 48.6 $ & $ 51.8 $ & $ 36.5 $ & \phantom{$-$}$ 5.28 $ & $ -2.18 $ & $ -4.90 $ & $ -3.35 $ \\
    \quad+Static & $ 55.2 $ & $ 40.4 $ & $ 49.0 $ & $ 52.0 $ & $ 33.5 $ & \phantom{$-$}$ 4.68 $ & $ \mathbf{-2.02} $ & $ -5.23 $ & $ -3.37 $ \\
    \quad+Adaptive & $ 53.7 $ & $ 41.0 $ & $ 49.0 $ & $ 52.1 $ & $ 34.2 $ & \phantom{$-$}$ 5.10 $ & $ -2.04 $ & $ -5.21 $ & $ -3.38 $ \\
    \quad+Randomize & $ 54.4 $ & $ 39.5 $ & $ 48.4 $ & $ 51.5 $ & $ 32.6 $ & \phantom{$-$}$ 3.81 $ & $ -2.08 $ & $ -5.16 $ & $ -3.38 $ \\
    Nucleus+$k$NN-MT & $ 51.6 $ & $ 42.1 $ & $ 50.4 $ & $ 52.8 $ & $ 37.0 $ & \phantom{$-$}$ \mathbf{14.5} $ & $ -2.37 $ & $ -4.50 $ & $ -3.33 $ \\
    \quad+Static & $ 55.0 $ & $ 42.7 $ & $ 49.9 $ & $ 52.5 $ & $ 34.9 $ & \phantom{$-$}$ 6.47 $ & $ -2.29 $ & $ -4.87 $ & $ -3.36 $ \\
    \quad+Adaptive & $ 55.6 $ & $ 42.6 $ & $ 49.8 $ & $ 52.4 $ & $ 34.7 $ & \phantom{$-$}$ 6.32 $ & $ -2.27 $ & $ -4.92 $ & $ -3.36 $ \\
    \quad+Randomize & $ \mathbf{59.4} $ & $ 42.3 $ & $ 49.2 $ & $ 52.0 $ & $ 33.1 $ & \phantom{$-$}$ 5.09 $ & $ -2.24 $ & $ -5.10 $ & $ -3.41 $ \\
     \bottomrule
  \end{tabular}
  \caption{Domain adaptation in German-English: We report averages of five domains.}\label{tab:dom_deen_average}
\end{table*}

\begin{table*}[htbp]
  \centering
  \small
  \begin{tabular}{@{}lcccccc|ccc@{}}
    \toprule
    & Diversity & \multicolumn{4}{c}{Translation Quality (BLEU $\uparrow$ )} & Both & \multicolumn{3}{c}{Fluency (PLL $\uparrow$ )} \\
    Method & DP $\uparrow$ & @1 & @20 & Merged@40 & Ref & DEQ $\uparrow$ & Max & Min & Mean \\
    \cmidrule[.05em](r){1-1} \cmidrule[.05em](lr){2-2} \cmidrule[.05em](lr){3-6} \cmidrule[.05em](lr){7-7} \cmidrule[.05em](l){8-10}
    Reference & - & - & - & - & - & - & - & - & $ -2.75 $ \\
    \cmidrule[.03em](r){1-1} \cmidrule[.03em](lr){2-2} \cmidrule[.03em](lr){3-6} \cmidrule[.03em](lr){7-7} \cmidrule[.03em](l){8-10}
    Baseline & $ 38.0 $ & $ 18.1 $ & $ 26.0 $ & $ 26.1 $ & $ 16.5 $ & $0.25$ & $ -1.75 $ & $ -3.67 $ & $ -2.55 $ \\
    DBS & $ 54.9 $ & $ 17.2 $ & $ 24.8 $ & $ 28.2 $ & $ 14.4 $ & $ 3.88 $ & $ -1.66 $ & $ -3.92 $ & $ -2.63 $ \\
    Nucleus & $ 63.9 $ & $ 17.6 $ & $ 26.5 $ & $ 28.9 $ & $ 14.8 $ & $ 6.43 $ & $ -1.69 $ & $ -3.72 $ & $ -2.60 $ \\
    $k$NN-MT & $ 37.4 $ & $ \mathbf{20.9} $ & $ \mathbf{29.7} $ & $ 31.5 $ & $ \mathbf{18.9} $ & - & $ -1.65 $ & $ \mathbf{-3.57} $ & $ \mathbf{-2.43} $ \\
    \cmidrule[.03em](r){1-1} \cmidrule[.03em](lr){2-2} \cmidrule[.03em](lr){3-6} \cmidrule[.03em](lr){7-7} \cmidrule[.03em](l){8-10}
    DBS+$k$NN-MT & $ 60.7 $ & $ 19.7 $ & $ 27.9 $ & $ 30.9 $ & $ 15.6 $ & $ 7.04 $ & $ -1.44 $ & $ -4.02 $ & $ -2.52 $ \\
    \quad+Static & $ 66.5 $ & $ 19.5 $ & $ 28.3 $ & $ 31.3 $ & $ 14.9 $ & $ 7.23 $ & $ -1.36 $ & $ -4.12 $ & $ -2.51 $ \\
    \quad+Adaptive & $ 66.8 $ & $ 19.5 $ & $ 28.4 $ & $ 31.3 $ & $ 14.8 $ & $ 7.18 $ & $ -1.36 $ & $ -4.11 $ & $ -2.51 $ \\
    \quad+Randomize & $ 65.9 $ & $ 19.2 $ & $ 27.8 $ & $ 30.9 $ & $ 14.6 $ & $ 6.61 $ & $ \mathbf{-1.35} $ & $ -4.13 $ & $ -2.51 $ \\
    Nucleus+$k$NN-MT & $ 66.6 $ & $ 20.3 $ & $ 29.0 $ & $ \mathbf{31.7} $ & $ 16.6 $ & $ 12.9 $ & $ -1.60 $ & $ -3.61 $ & $ -2.49 $ \\
    \quad+Static & $ 64.0 $ & $ 20.5 $ & $ 28.9 $ & $ 31.5 $ & $ 16.8 $ & $ 12.9 $ & $ -1.56 $ & $ -3.65 $ & $ -2.46 $ \\
    \quad+Adaptive & $ 64.0 $ & $ 20.6 $ & $ 28.8 $ & $ 31.5 $ & $ 16.9 $ & $ \mathbf{13.2} $ & $ -1.57 $ & $ -3.65 $ & $ -2.46 $ \\
    \quad+Randomize & $ \mathbf{74.8} $ & $ 20.3 $ & $ 28.7 $ & $ 31.4 $ & $ 14.6 $ & $ 8.71 $ & $ -1.50 $ & $ -4.16 $ & $ -2.59 $ \\
    \bottomrule
  \end{tabular}
  \caption{Domain adaptation in Japanese-English: We report averages of four domains.}\label{tab:dom_jaen_average}
\end{table*}

\subsubsection{Domain Adaptation}
Summaries of the De-En and Ja-En results are shown in Table~\ref{tab:dom_deen_average} and Table~\ref{tab:dom_jaen_average}, respectively, by averaging the metrics across the domains.
Detailed results are shown in Appendix~\ref{appendix:detailed_results}.

In De-En, our proposed DBS+$k$NN-MT and Nucleus+$k$NN-MT outperformed both of DP and oracle BLEU of DBS and Nucleus.
These methods also decrease BLEU@20 more than $k$NN-MT, although the drop in performance is comparable to that observed between Baseline and DBS or Nucleus.
Our perturbation methods, i.e., +Adaptive, +Static, and +Randomize, drastically improved DP while maintaining comparable performance to DBS+ and Nucleus+$k$NN-MT under BLEU@20.
Nor did the PLL of the proposed methods suffer substantial drops when compared to $k$NN-MT; the differences are marginal compared to the PLL of Reference.
In Ja-En, the proposed methods improved DP like in De-En without any MergedBLEU@40 loss.\footnote{We also evaluate the oracle BLEU for 40 candidates~(BLEU@40) to compare to MergedBLEU@40, and the results are in Appendix~\ref{appendix:ablation_for_merged_oracle}.}
The PLL of the proposed methods is also comparable to Baseline.

We observed almost no substantial differences for the perturbation types.
+Static requires prior estimation on distance metrics, and +Adaptive needs an additional $k$NN search for each time step for the inferences.
Therefore, +Randomize is the best choice since it overcomes both drawbacks.

These results indicate that the proposed methods improved the diversity without lowering the fluency and maintained oracle translation quality on some domains.

\paragraph{Trade-off between quality and diversity}
We observed our proposed methods suffered from a quality-diversity trade-off~\cite{ippolito-etal-2019-comparison, zhang-etal-2021-trading}, i.e., our methods improve diversity (DP) but decrease average translation quality (RefBLEU).
However, all of our proposed DBS- and Nucleus-based methods outperformed the DEQ of DBS and Nucleus.\footnote{$k$NN-MT is not comparable to our methods because the base of the DEQ is $k$NN-MT.}
Thus, our methods achieved better quality-diversity trade-offs than the existing methods.

\begin{table*}[tbp]
  \centering
  \small
  \begin{tabular}{@{}lcccccc|ccc@{}}
    \toprule
    & Diversity & \multicolumn{4}{c}{Translation Quality (BLEU $\uparrow$ )} & Both & \multicolumn{3}{c}{Fluency (PLL $\uparrow$ )} \\
    Method & DP $\uparrow$ & @1 & @20 & Merged@40 & Ref & DEQ $\uparrow$ & Max & Min & Mean \\
    \cmidrule[.05em](r){1-1} \cmidrule[.05em](lr){2-2} \cmidrule[.05em](lr){3-6} \cmidrule[.05em](lr){7-7} \cmidrule[.05em](l){8-10}
    Reference & - & - & - & - & - & - & - & - & $ -2.83 $ \\
    \cmidrule[.03em](r){1-1} \cmidrule[.03em](lr){2-2} \cmidrule[.03em](lr){3-6} \cmidrule[.03em](lr){7-7} \cmidrule[.03em](l){8-10}
    Baseline & $ 37.5 $ & $ 30.2 $ & $ 41.2 $ & $ 41.2 $ & $ 27.6 $ & $1.24$ & $ -1.94 $ & $ -3.85 $ & $ -2.77 $ \\
    DBS & $ 51.3 $ & $ 28.9 $ & $ 37.9 $ & $ 43.1 $ & $ 24.3 $ & $ 3.97 $ & $ -1.78 $ & $ -4.04 $ & $ -2.78 $ \\
    Nucleus & $ 62.8 $ & $ 29.2 $ & $ 40.4 $ & $ 44.2 $ & $ 23.7 $ & $ 6.16 $ & $ -1.78 $ & $ -3.99 $ & $ -2.76 $ \\
    $k$NN-MT & $ 37.3 $ & $ \mathbf{30.5} $ & $ \mathbf{41.4} $ & $ 42.6 $ & $ \mathbf{27.8} $ & - & $ -1.92 $ & $ -3.85 $ & $ -2.76 $ \\
    \cmidrule[.03em](r){1-1} \cmidrule[.03em](lr){2-2} \cmidrule[.03em](lr){3-6} \cmidrule[.03em](lr){7-7} \cmidrule[.03em](l){8-10}
    DBS+$k$NN-MT & $ 52.6 $ & $ 29.1 $ & $ 38.2 $ & $ 43.4 $ & $ 24.2 $ & $ 4.22 $ & $ -1.70 $ & $ -4.09 $ & $ -2.76 $ \\
    \quad+Static & $ 54.8 $ & $ 29.0 $ & $ 38.3 $ & $ 43.4 $ & $ 23.8 $ & $ 4.38 $ & $ -1.66 $ & $ -4.16 $ & $ -2.75 $ \\
    \quad+Adaptive & $ 54.3 $ & $ 29.0 $ & $ 38.3 $ & $ 43.4 $ & $ 23.9 $ & $ 4.34 $ & $ -1.67 $ & $ -4.15 $ & $ -2.75 $ \\
    \quad+Randomize & $ 53.9 $ & $ 29.0 $ & $ 38.2 $ & $ 43.4 $ & $ 23.9 $ & $ 4.29 $ & $ -1.66 $ & $ -4.14 $ & $ -2.75 $ \\
    \quad+Uniquify & $ 54.9 $ & $ 28.8 $ & $ 37.8 $ & $ 43.2 $ & $ 23.5 $ & $ 4.08 $ & $ -1.69 $ & $ -4.18 $ & $ -2.78 $ \\
    \quad\quad+Static & $ 55.9 $ & $ 28.8 $ & $ 37.7 $ & $ 43.1 $ & $ 23.3 $ & $ 4.09 $ & $ -1.65 $ & $ -4.24 $ & $ -2.77 $ \\
    \quad\quad+Adaptive & $ 55.7 $ & $ 28.8 $ & $ 37.8 $ & $ 43.2 $ & $ 23.3 $ & $ 4.10 $ & $ -1.65 $ & $ -4.22 $ & $ -2.78 $ \\
    \quad\quad+Randomize & $ 55.8 $ & $ 28.8 $ & $ 37.7 $ & $ 43.2 $ & $ 23.3 $ & $ 4.09 $ & $ \mathbf{-1.64} $ & $ -4.25 $ & $ -2.78 $ \\
    \cmidrule[.03em](r){1-1} \cmidrule[.03em](lr){2-2} \cmidrule[.03em](lr){3-6} \cmidrule[.03em](lr){7-7} \cmidrule[.03em](l){8-10}
    Nucleus+$k$NN-MT & $ 64.5 $ & $ 29.1 $ & $ 40.5 $ & $ \mathbf{44.3} $ & $ 23.5 $ & $ 6.22 $ & $ -1.73 $ & $ -4.02 $ & $ -2.75 $ \\
    \quad+Static & $ 52.4 $ & $ 30.2 $ & $ 38.9 $ & $ 43.4 $ & $ 25.8 $ & $ \mathbf{7.25} $ & $ -1.85 $ & $ \mathbf{-3.80} $ & $ \mathbf{-2.73} $ \\
    \quad+Adaptive & $ 52.8 $ & $ 30.0 $ & $ 38.9 $ & $ 43.4 $ & $ 25.5 $ & $ 6.75 $ & $ -1.86 $ & $ -3.83 $ & $ -2.74 $ \\
    \quad+Randomize & $ 62.3 $ & $ 29.8 $ & $ 38.9 $ & $ 43.5 $ & $ 23.4 $ & $ 5.64 $ & $ -1.76 $ & $ -4.13 $ & $ -2.77 $ \\
    \quad+Uniquify & $ \mathbf{70.8} $ & $ 28.0 $ & $ 39.3 $ & $ 44.0 $ & $ 21.0 $ & $ 4.88 $ & $ -1.70 $ & $ -4.16 $ & $ -2.78 $ \\
    \quad\quad+Static & $ 55.4 $ & $ 29.8 $ & $ 38.7 $ & $ 43.3 $ & $ 24.9 $ & $ 6.09 $ & $ -1.84 $ & $ -3.88 $ & $ -2.75 $ \\
    \quad\quad+Adaptive & $ 55.7 $ & $ 29.9 $ & $ 38.7 $ & $ 43.4 $ & $ 24.8 $ & $ 6.05 $ & $ -1.84 $ & $ -3.92 $ & $ -2.76 $ \\
    \quad\quad+Randomize & $ 67.7 $ & $ 29.5 $ & $ 38.4 $ & $ 43.3 $ & $ 21.6 $ & $ 4.86 $ & $ -1.75 $ & $ -4.44 $ & $ -2.84 $ \\
    \bottomrule
  \end{tabular}
  \caption{General domain: We report averages of three language pairs.}\label{tab:general}
\end{table*}

\subsubsection{General domain}
Table~\ref{tab:general} summarizes the general-domain results obtained by averaging the metrics across the language pairs.
Detailed results are shown in Appendix~\ref{appendix:detailed_results}.

The proposed DBS+ and Nucleus+$k$NN-MT slightly improved DP, and the MergedBLEU@40 and fluency are comparable to DBS and Nucleus.
The effect of stochastic perturbations for DP was limited, especially on Nucleus-based, but +Uniquify substantially improved DP, and MergedBLEU@40 and fluency preserved comparable results.
As in the domain-adaptation setting, the DEQ of our methods outperformed existing methods.

These experiments show that the proposed methods achieve better quality-diversity trade-offs without any fluency loss.

\section{Analysis}
\subsection{\texorpdfstring{Tuning $k$NN Diversified Decoding}{}}
\label{sec:analysis_tradeoff}
\begin{figure}[tbp]
    \centering
    \includegraphics[width=0.9\linewidth]{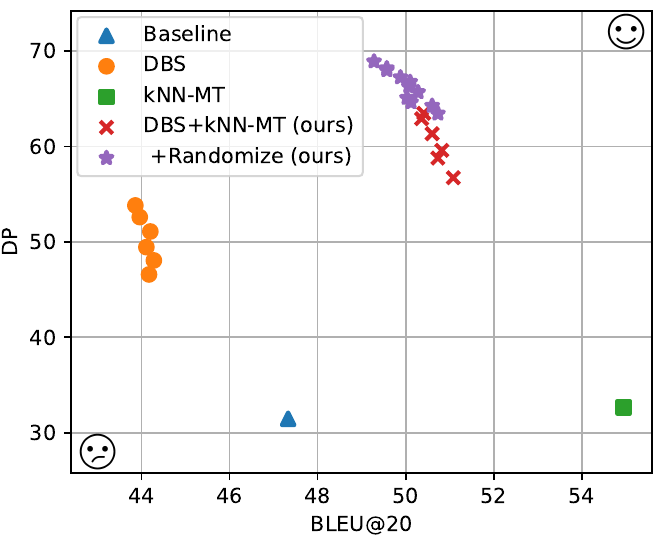}
    \caption{Relationship between translation quality~(BLEU@20) and diversity (DP) in De-En IT domain: Top-right is most desirable.}
    \label{fig:tradeoff}
\end{figure}

We investigated how the hyperparameters of our proposed method affect its performance.
Figure~\ref{fig:tradeoff} shows the relationship between DP and BLEU@20 in the De-En IT domain.\footnote{For DBS and DBS+$k$NN-MT, we varied DBS's diversity strength by $0.1$ in the range of $\lbrack 1.5, 2.0 \rbrack$.
For +Randomize, we used $1.5$ for diversity strength and varied perturbation's magnitude $h$ by $0.1$ in the range of $\lbrack 1.5, 2.5 \rbrack$.}
The results show that +Randomize outperformed the diversity of DBS+$k$NN-MT while maintaining oracle translation quality with some hyperparameters, indicating that our proposed method can adjust DP and BLEU by varying the magnitude of the perturbation.\footnote{Further analysis of the relationship between the hyperparameters and DP/BLEU of our methods is in Appendix~\ref{appendix:tuning_proposed_method}.}

\subsection{Overcorrection Analysis}
\label{sec:overcorrection_analysis}
We hypothesized that the overcorrection problem discourages the generation of diverse candidates that is alleviated by our proposed methods.
To verify the hypothesis, we evaluated how well our methods mitigate the overcorrection problem and clarified the relationship between overcorrection and diversity. 

Overcorrection is a phenomenon in which the likelihoods of valid translations are underestimated by a model.
Therefore, a model that suffers less from the issue will assign a similar likelihood to valid translations that only have small differences.
Thus, we propose a mean of the absolute difference in the log-likelihoods (MADLL) of two reference translations as a metric that quantifies the degree of overcorrection, in which a lower MADLL value implies a decreased likely of suffering from overcorrection issue.\footnote{We report MADLL along with BLEU because it is easy to improve only MADLL but hard to improve both (if the model assigns the same likelihood to all sentences, MADLL will be zero, but BLEU will be substantially affected).}
We evaluated the proposed methods on the test data of WMT'21 De-En (newstest2021) and WMT'22 De-En (generaltest2022) in the De-En general-domain setting. 
These test data have two reference translations (refA/refB) for one source sentence, and we report the MADLL between refA and refB by forced decoding. 

Table~\ref{tab:overcorrection} shows the relationship between overcorrection, diversity, and translation quality.
The proposed methods have lower MADLL and higher DP scores than Baseline and $k$NN-MT for both WMT'21 and WMT'22.
We also found that BLEUs of the Baseline and $k$NN-MT are almost comparable to the proposed methods. 
This implies that the proposed methods managed to resolve overcorrection and improved diversity while almost maintaining the translation quality.
\begin{table}[tbp]
  \centering
  \small
  \tabcolsep 3.4pt
  \begin{tabular}{@{}lcccc@{}}
  \toprule
     & MADLL $\downarrow$ & DP $\uparrow$ & BLEU@1 $\uparrow$ & BLEU@20 $\uparrow$ \\ \midrule
    Method & \multicolumn{4}{c}{WMT'21 (newstest2021)} \\ \cmidrule(r){1-1} \cmidrule(l){2-5}
    Baseline & $ 0.695 $ & $ 41.0 $ & $ 29.5 $ / $ 36.0 $ & $38.2$ / $45.4$ \\
    $k$NN-MT & $0.683$ & $41.1$ & $\mathbf{30.1}$ / $\mathbf{36.8}$ & $\mathbf{38.7}$ / $\textbf{45.8}$ \\
    \quad+Uniq & $\mathbf{0.657}$ & $43.6$ & $29.6$ / $36.2$ & $38.0$ / $45.5$ \\
    \quad\quad+Rand & $0.660$ & $\mathbf{43.9}$ & $29.6$ / $36.3$ & $37.8$ / $44.7$ \\ \midrule
    Method & \multicolumn{4}{c}{WMT'22 (generaltest2022)} \\ \cmidrule(r){1-1} \cmidrule(l){2-5}
    Baseline & $0.712$ & $42.2$ & $30.3$ / $34.5$ & $38.3$ / $43.1$ \\
    $k$NN-MT & $0.714$ & $42.3$ & $\mathbf{30.6}$ / $\mathbf{35.0}$ & $\mathbf{38.6}$ / $\mathbf{43.5}$ \\
    \quad+Uniq & $\mathbf{0.696}$ & $44.9$ & $30.5$ / $34.7$ & $38.2$ / $42.9$ \\
    \quad\quad+Rand & $0.702$ & $\mathbf{45.2}$ & $30.5$ / $34.7$ & $38.0$ / $42.5$ \\
  \bottomrule
  \end{tabular}
  \caption{Overcorrection analysis on newstest2021 and generaltest2022 in De-En: MADLL is an indicator where a lower score denotes less likely to suffer from overcorrection. DP and BLEUs are scores when DBS is used as the decoding method. BLEU is written in the form of scores for refA/refB. Uniq and Rand are abbreviations for Uniquify and Randomize, respectively.}\label{tab:overcorrection}
\end{table}

\begin{figure*}[tbp]
\centering
\small
\begin{tabular}{@{}p{7.6cm}p{7.6cm}@{}}
        \toprule[1.5pt]
\mbox{\textbf{Test Input}: \emph{\Ja{コロナに関しまして。}}}          & \\
\mbox{\textbf{Reference}: \emph{I have a question about COVID.}} & \\
                     \cmidrule[0.75pt](){1-2}
 \textbf{DBS+$k$NN-MT+Randomize} & \textbf{DBS+$k$NN-MT} \\ 
                     \cmidrule[0.75pt](){1-2}
\emph{About corona.} & \emph{Regarding corona.} \\ \addlinespace[0.1em]
\emph{With regards to corona.} & \emph{About corona.} \\ \addlinespace[0.1em]
\emph{About \underline{COVID-19}.} & \emph{It is about corona.} \\ \addlinespace[0.1em]
\emph{Regarding corona.} & \emph{We are talking about corona.} \\ \addlinespace[0.1em]
\emph{With regards to \underline{COVID-19}.} & \emph{With regards to corona.} \\
\centering{{\tiny \Ja{⁝}}} & \centering{{\tiny \Ja{⁝}}} \arraybackslash \\
\cmidrule[1.00pt](){1-2}
\mbox{\textbf{Test Input}: \emph{\Ja{Spring Summerコレクションもセール対象商品!}}}          & \\
\mbox{\textbf{Reference}: \emph{The spring/summer collection is also included in the sale!}} & \\
                     \cmidrule[0.75pt](){1-2}
\textbf{DBS+$k$NN-MT+Randomize} & \textbf{DBS} \\ 
                     \cmidrule[0.75pt](){1-2}
\emph{The Spring Summer collection is also a sale target product!} & \emph{The Spring Summer collection is also a sale target product!} \\ \addlinespace[0.1em]
\emph{Items from the Spring Summer collection are also on sale!} & \emph{The Spring Summer collection is also a sale item!} \\ \addlinespace[0.1em]
\emph{The Spring Summer collection is also a sale target product!} & \emph{The Spring Summer collection is also a sale eligible product!} \\
\centering{{\tiny \Ja{⁝}}} & \centering{{\tiny \Ja{⁝}}} \arraybackslash \\
\emph{\underline{winter} collection is also a sale target product!} & \emph{Our Spring Summer collections are on sale!} \\ \addlinespace[0.1em]
\emph{The Spring Summer collection is also eligible for sale.} & \emph{The Spring Summer collection is also eligible for sale!} \\ \addlinespace[0.1em]
\emph{Summer collection is also a sale target product!} & \emph{The Spring Summer Collection is also included in the sale!} \\ \addlinespace[0.1em]
\bottomrule[1.5pt]
\end{tabular}
    \caption{Example 20-best lists using DBS-based methods: In upper example, DBS+$k$NN-MT+Randomize successfully diversified list by adding a likely word, \emph{COVID-19}, which did not appear in DBS+$k$NN-MT. In lower example, DBS+$k$NN-MT+Randomize introduced an unlikely word, \emph{winter}, which did not appear in DBS.}
    \label{fig:ex1}
\end{figure*}

\subsection{\texorpdfstring{Counting Distinct $n$-grams}{}}
\label{sec:ngram_types}
In \S\ref{sec:experiments}, we used DP as a diversity metric.
DP captures how many unique $n$-grams are included in each candidate.
In order to evaluate the diversity of translation candidates of our proposed methods from a different perspective, we employed another metric: the number of distinct $n$-grams, which measures the richness of vocabulary and phrases across the entire $N$-best list.
We calculated the ratio of the number of distinct $n$-grams to the total number of $n$-grams for $n \in \{1,2,3,4\}$.

The ratio averages in the De-En domain adaptation setting are shown in Table~\ref{tab:n-gram}.
DBS+ and Nucleus+$k$NN-MT increased the ratio of the number of distinct $n$-grams more than DBS and Nucleus; the ratio increased substantially when perturbation was applied to it.
The results show that our proposed methods generate translation candidates with more diverse vocabulary and phrases compared to the baselines.

\begin{table}[tbp]
  \centering
  \small
  \begin{tabular}{@{}lcccc@{}}
  \toprule
     & \multicolumn{4}{c}{Ratio of distinct $n$-grams (\%) $\uparrow$} \\
    Method & $n=1$ & $n=2$ & $n=3$ & $n=4$  \\
    \cmidrule[.05em](r){1-1} \cmidrule[.05em](l){2-5}
    Baseline & $ 1.6 $ & \phantom{$0$}$ 7.8 $ & $ 14.1 $ & $ 18.5 $ \\ \cmidrule(r){1-1} \cmidrule(l){2-5}
    DBS & $ 1.6 $ & \phantom{$0$}$ 8.8 $ & $ 16.9 $ & $ 22.2 $ \\
    DBS+$k$NN-MT & $ 1.7 $ & \phantom{$0$}$ 9.7 $ & $ 18.6 $ & $ 24.8 $ \\
    \quad+Randomize & $ 2.1 $ & $ 12.4 $ & $ 24.7 $ & $ 33.4 $ \\  \cmidrule(r){1-1} \cmidrule(l){2-5}
    Nucleus & $ 2.0 $ & $ 11.1 $ & $ 23.1 $ & $ 32.8 $ \\
    Nucleus+$k$NN-MT & $ 1.9 $ & $ 11.9 $ & $ 25.9 $ & $ 37.1 $ \\
    \quad+Randomize & $ \mathbf{2.6} $ & $ \mathbf{16.0} $ & $ \mathbf{32.2} $ & $ \mathbf{42.8} $ \\
  \bottomrule
  \end{tabular}
  \caption{The ratio of the number of distinct $n$-grams to the total number of $n$-grams in German-English domain adaptation setting: We report averages of five domains.}\label{tab:n-gram}
\end{table}

\begin{table}[tbp]
  \centering
  \small
  \tabcolsep 3.3pt
  \begin{tabular}{@{}ccc|ccc@{}}
    \toprule
    \#neighbors & $h$ & $\lfloor h \times k \rfloor$ & DP $\uparrow$ & BLEU@1 $\uparrow$ & BLEU@20 $\uparrow$ \\ \midrule
    \multicolumn{6}{c}{(1) DBS+$k$NN-MT+Randomize} \\ 
    \cmidrule{1-6}
    $64$ & $2$ & $128$ & $57.1$ & $41.5$ & $50.8$ \\
    $64$ & $3$ & $192$ & $62.3$ & $39.3$ & $50.0$ \\
    $64$ & $4$ & $256$ & $\mathbf{65.6}$ & $38.1$ & $49.3$ \\
    \midrule
    \multicolumn{6}{c}{(2) DBS+$k$NN-MT} \\
    \cmidrule{1-6}
    $128$ & - & - & $44.2$ & $\mathbf{43.9}$ & $51.1$ \\
    $192$ & - & - & $44.1$ & $\mathbf{43.9}$ & $\mathbf{51.3}$ \\
    $256$ & - & - & $44.2$ & $\mathbf{43.9}$ & $51.1$ \\
    \bottomrule
  \end{tabular}
  \caption{Effectiveness of \emph{Randomize} on De-En IT domain based on DBS+$k$NN-MT: We compared (1) randomize $k$ from $\lfloor h \times k \rfloor$ neighbors and (2) set number of neighbors per query to $\lfloor h \times k \rfloor$.}\label{tab:ablation_randomize}
\end{table}

\subsection{Effectiveness of Randomization}
We conducted an ablation study to investigate the effectiveness of \emph{Randomize} on the Randomized-$k$NN.
In the Randomized-$k$NN, the search space is stochastically expanded by uniformly and randomly sampling $k$ from $\lfloor h \times k \rfloor$ neighbors to diversify the translations.
We compared the following two methods to investigate the effectiveness of \emph{Randomize}:
(1) randomizing $k$ from $\lfloor h \times k \rfloor$ neighbors with DBS+$k$NN-MT, i.e., DBS+$k$NN-MT+Randomize, and (2) retrieving $\lfloor h \times k \rfloor$ neighbors without randomizing on DBS+$k$NN-MT i.e., setting the number of neighbors per query of DBS+$k$NN-MT to $\lfloor h \times k \rfloor$.

A comparison for the De-En IT domain is presented in Table~\ref{tab:ablation_randomize}, where simply increasing the number of neighbors per query of DBS+$k$NN-MT did not improve diversity.
\emph{Randomize} from more neighbors is important for improving diversity.

\subsection{Case Study}
To better understand our proposed method through case studies, Figure~\ref{fig:ex1} shows two qualitative examples in the general domain of Ja-En.
We omitted some parts for brevity, and a full version is shown in Figure~\ref{fig:ex_full} in Appendix~\ref{sec:appendix_quantitative}.

In the upper example, Randomized-$k$NN improved the diversity of the candidates, which include the appropriate word \emph{COVID-19}.
This candidate never appeared in the 20-best list generated by DBS+$k$NN-MT, suggesting that considering more likely tokens by +Randomize with a broader search space improves diversity and maintains translation quality.

The example at the bottom shows increased diversity but also decreased translation quality, where translation \emph{winter} is output for \emph{spring/summer}, which does not appear in the DBS-generated candidates.
Such antonyms as \emph{winter}, \emph{spring}, and \emph{summer} tend to appear in the neighbors of word embedding space~\citep{mrksic-etal-2016-counter}, which is the primary cause of incorrect retrieval from the datastore in the broader $k$NN search space.
We leave it as our future work of addressing the problem of retrieving unlikely words by a stochastically expanded $k$NN search.

\section{Conclusion}
We proposed methods to generate more diverse translation candidates by expanding the search space of $k$NN-MT.
We experimentally showed that our proposed methods alleviated the overcorrection problem and outperformed the existing baselines in diversity, and also controlled the diversity and translation quality by changing the perturbation's magnitude.

\section*{Limitations}
Our proposed method improves diversity by utilizing $k$NN-MT.
Unfortunately, $k$NN-MT suffers from the drawbacks of high inference latency for $k$NN searches and requires much memory to load the datastore.
Our proposed method is applicable not only to vanilla-$k$NN but also to many other variants; if a model is proposed in the future that solves these issues, we can combine our method with new $k$NN-MT variants to overcome these drawbacks.

Although our proposed method improves diversity, it might generate hallucinations, which are incorrect but fluent translations.
This problem can be alleviated by filtering hallucinations by post-processing, an approach we leave for the future.

We also might need to consider the trade-off between diversity and quality depending on downstream applications, as in a number of experiments.

We showed the effectiveness of our proposed methods by evaluating the diversity and oracle translation quality, but the benefit in end-applications remains unclear.
\citet{li2016mutual} implied that the higher diversity of translation candidates promotes the higher translation quality after reranking.
Thus, the benefit in downstream applications can be shown by measuring the performance after using a reranking method such as quality-aware decoding~\cite{fernandes-etal-2022-quality}.

\section*{Acknowledgements}
This work was supported by JSPS KAKENHI Grant Numbers JP21H05054, JP21K17801, and JP23H03458.

\bibliography{anthology,custom}

\clearpage
\appendix
\section{Detailed Experimental Settings}
\subsection{Statistics of Dataset}
\label{appendix:detailed_dataset}
Table~\ref{tab:dataset} shows the dataset's statistics. 
$|\mathcal{D}|$ is the size of the datastore (identical to the number of target-side tokens of the training data).
\#train and \#test are the number of sentences in the training and the test data.

\begin{table}[tbp]
  \centering
  \small
  \begin{tabular}{@{}lc|rcr@{}}
  \toprule
    Corpus & Src-Tgt & \multicolumn{1}{c}{\#train} & $|\mathcal{D}|$ & \multicolumn{1}{c}{\#test} \\ \midrule
    \multicolumn{5}{c}{Domain Adaptation} \\ \cmidrule{1-5}
    Koran & \multirow{5}{*}{De-En} & 14,979 & 450K & 2,000 \\
    IT &  & 177,795 & 3.10M & 2,000 \\
    Medical &  & 206,804 & 5.70M & 2,000 \\
    Law &  & 447,701 & 18.4M & 2,000 \\
    Subtitles &  & 12,409,630 & 154M & 2,000 \\ \cmidrule{1-5}
    ASPEC & \multirow{4}{*}{Ja-En} & 2,000,000 & 68.3M & 1,812 \\
    KFTT &  & 440,288 & 15.2M & 1,160 \\
    TED talk &  & 223,108 & 5.24M & 1,285 \\
    BSD &  & 20,000 & 256K & 2,120 \\ \midrule
    \multicolumn{5}{c}{General Domain} \\ \cmidrule{1-5}
    WMT'19 & De-En & 32,278,623 & 916M & 2,000 \\
    WMT'22 & Ja-En & 32,104,268 & 874M & 2,008 \\
    WMT'22 & Uk-Cs & 12,621,881 & 192M & 2,812 \\
  \bottomrule
  \end{tabular}
  \caption{Statistics of dataset}\label{tab:dataset}
\end{table}

\subsection{Model Settings}
\label{appendix:detailed_models}
Table~\ref{tab:hyperparameter} shows the hyperparameters we used in the experiments.

\paragraph{Nucleus sampling}
We tuned hyperparameter $p$ from $p \in \{0.1, 0.2, 0.3, 0.4, 0.5,\allowbreak 0.6,\allowbreak 0.7,\allowbreak 0.8,\allowbreak 0.9,\allowbreak 0.95\}$ based on the validation data.

\paragraph{$k$NN-MT}
We used squared-L2 distance as a distance function.
For efficiency, we quantized the datastore with IVFPQ and we set the code size to 64.
We used the 1024-dimensional representation input to the final layer feedforward network as the key.
For the domain adaptation settings, we used 1M keys with 4096 clusters.
For the general-domain settings, we used 5M keys with 65536 clusters.
For inference, neighbors were searched from the nearest 32 clusters in the datastore.
For the De-En domain adaptation setting, we used the same settings as \newcite{khandelwal2021nearest} for $k$, $\lambda$, and $\tau$.
For the Ja-En domain adaptation setting, we used the same $k$ as \newcite{khandelwal2021nearest} and tuned $\lambda$ and $\tau$ from $\lambda \in \{0.1,0.2,...,0.9\}$, $\tau \in \{10, 100, 1000\}$ with validation data.
For the general-domain settings, we tuned hyperparameters $k$, $\lambda$, and $\tau$ from $k \in \{16,32,64,128\}$, $\lambda \in \{0.1,0.2,...,0.9\}$, $\tau \in \{10, 100, 1000\}$ with validation data.

\paragraph{Proposed method}
For the DBS+* settings, we used the same parameters as the baseline.
We tuned $p$ from $p \in \{0.1,0.2,0.3,0.4,0.5,0.6,0.7,0.8,0.9,0.95\}$ for Nucleus+* settings without +Perturbation and from $p \in \{0.1,0.3,0.5,0.7,0.9\}$ for setting with +Perturbation.
For the +Static settings, we computed the mean $d_{m}$ and standard deviation $d_{s}$ of the distance to the nearest neighbors on the validation data in advance, and set $h_{m}=h'_{m} \times d_{m}$ and $h_{s}=h'_{s} \times d_{s}$, where $h'_{m}$ and $h'_{s}$ are tuned parameters from $h'_{m} \in \{0.025,0.05,0.1,0.2,0.4,0.8\}$ and $h'_{s} \in \{0.025,0.05,0.1,0.2,0.4,0.8\}$ on the validation data.
For the +Adaptive settings, we tuned the hyperparameter from $h'_{m} \in \{0.025,0.05,0.1,0.2,0.4,0.8\}$, $h'_{s} \in \{0.025,0.05,0.1,0.2,0.4,0.8\}$.
For the +Randomize settings, we tuned the hyperparameters from $h \in \{1.1,1.2,...,4.0\}$.
Note that the hyperparameters for +$k$NN-MT, such as $k$, $\lambda$, and $\tau$, on the proposed methods are identical to the standard $k$NN-MT.

\subsection{Metric Settings}
\label{appendix:detailed_metrics}
The detailed metric settings are as follows:

\paragraph{BLEU}
is calculated with \texttt{sacrebleu}~\cite{post-2018-call}.
The signature for the corpus-wise BLEU is \texttt{nrefs:1|case:mixed|eff:no\\|tok:13a|smooth:none|version:2.2.1}, and for the sentence-level BLEU is \texttt{nrefs:1|case:mixed|eff:yes|tok:13a\\|smooth:add-k[1.00]|version:2.2.1}.

\paragraph{MedBLEU}
is the corpus-wise BLEU score computed by the median sentence-level BLEU score for each $N$-best candidates.
When $N$ is even, we selected the sentence with the highest sentence-level BLEU between the two sentences in the middle.

\paragraph{PLL}
is a metric of the fluency and is computed for sentence $\boldsymbol{y} = (w_{1}, \hdots, w_{|\boldsymbol{y}|})$:
\begin{align}
\label{eq:PLL}
    \text{PLL}(\boldsymbol{y})=\sum_{t=1}^{|\boldsymbol{y}|} \log P_{\text{MLM}}(w_{t}|\boldsymbol{y}_{\backslash t}),
\end{align}
where $\boldsymbol{y}_{\backslash t}$ is a sentence with masked token $w_{t}$ at time step $t$ and $P_{\text{MLM}}(w_{t}|\boldsymbol{y}_{\backslash t})$ is the probability that the MLM model predicts original token $w_{t}$ from masked sentence $\boldsymbol{y}_{\backslash t}$.

\paragraph{DP}
is formally defined for $N$-best candidates for source sentence set $\mathcal{X}$ as $\mathbb{H}=\{ \mathbf{H}_{1}, \hdots, \mathbf{H}_{N} \}$, where $\mathbf{H}_{n}=\{ \hat{\boldsymbol{y}}^{n}_{1}, \hdots, \hat{\boldsymbol{y}}^{n}_{M} \}$, is calculated as follows:
\begin{multline}
\label{eq:DP}
    \text{DP}(\mathbb{H})= \frac{1}{N(N-1)} \\
    \times \sum_{\mathbf{H} \in \mathbb{H}} \sum_{\mathbf{H}' \in \mathbb{H}, \mathbf{H}' \neq \mathbf{H}} 1-\text{BLEU}(\mathbf{H}, \mathbf{H}').
\end{multline}
Note that $\text{BLEU}(\mathbf{H}, \mathbf{H}')$ is the corpus-wise BLEU of hypothesis $\mathbf{H}$ for reference $\mathbf{H}'$.

We also used the following metrics to further evaluate the proposed methods in detail.
The results are in Appendix~\ref{appendix:detailed_results}.

\paragraph{MeanLen}
is the mean sentence length ratio of the candidates to the reference translations.
The closer this metric is to $1$, indicating that the model outputs sentences of more appropriate length.

\paragraph{COMET@N}
is the system-level COMET~\citep{rei-etal-2020-comet} score computed by the largest sentence-level COMET score for each $N$-best candidates.
We use \texttt{wmt22-comet-da} model\footnote{\url{https://huggingface.co/Unbabel/wmt22-comet-da}} for evaluation, and report COMET@1 and COMET@20 in our experiment.

\paragraph{BERTScore@N}
is the system-level BERTScore~\citep{Zhang*2020BERTScore:} computed by the largest sentence-level BERTScore for each $N$-best candidates.
We report BERTScore@1 and BERTScore@20 in our experiment.
The hashcode for BERTScore is \texttt{roberta-large\_L17\_idf\_version=0.3.12\\(hug\_trans=4.22.2)-rescaled}.

\paragraph{Speed}
is the inference speed (tokens/s) logged by \texttt{fairseq} when using a single GPU (GeForce RTX 3090).

\section{Detailed Results}
\label{appendix:detailed_results}
The results for each domain of the De-En domain adaptation setting are shown in Table~\ref{tab:koran} to Table~\ref{tab:subtitles}.
The results for each domain of the Ja-En domain adaptation setting are shown in Table~\ref{tab:aspec} to Table~\ref{tab:bsd}.
The results for each language pair of the general-domain setting are shown in Table~\ref{tab:general_deen} to Table~\ref{tab:general_ukcs}.

\section{Further Analysis}

\subsection{Ablation Study for MergedBLEU@N}
\label{appendix:ablation_for_merged_oracle}
In \S \ref{sec:experiments}, we evaluated MergedBLEU@40, the oracle translation quality when merged with Baseline, and showed that the proposed methods' MergedBLEU@40 are comparable to baselines (Baseline, DBS, Nucleus and $k$NN-MT) in the Ja-En domain adaptation and general-domain settings.
However, it is not obvious whether the proposed methods' MergedBLEU@40 is also comparable to the oracle quality of baselines with a larger beam size.
Thus, we conducted an ablation study.

Tables \ref{tab:oracle40_de-en}, \ref{tab:oracle40_ja-en}, and \ref{tab:oracle40_general} show the oracle BLEU results for the 40-best (BLEU@40) when the baselines' beam size are set to 40.\footnote{For evaluating BLEU@40, we used the same hyperparameters as in \S \ref{sec:experiments}
 and Appendix~\ref{appendix:detailed_models} except for beam size.}
We found that the MergedBLEU@40 of our proposed methods even shows comparable performance to BLEU@40 of baselines in the Ja-En domain adaptation (Table~\ref{tab:oracle40_ja-en}) and general-domain (Table~\ref{tab:oracle40_general}) settings.
These results support our hypothesis that our diversified methods generate high-quality candidates.

\begin{table}[tbp]
  \centering
  \small
  \begin{tabular}{@{}lcccc@{}}
    \toprule
    & \multirow{2}{*}{DP} & \multicolumn{3}{c}{BLEU} \\
    Method & & @20 & @40 & Mrg@40 \\
    \cmidrule[.05em](r){1-1} \cmidrule[.05em](lr){2-2} \cmidrule[.05em](l){3-5}
    Baseline & $ 31.4 $ & $ 42.6 $ & $ 44.4 $ & - \\
    DBS & $ 35.9 $ & $ 40.0 $ & $ 41.6 $ & $ 43.8 $ \\
    Nucleus & $ 48.0 $ & $ 42.1 $ & $ 43.7 $ & $ 44.6 $ \\
    $k$NN-MT & $ 32.3 $ & $ \mathbf{51.8} $ & $ \mathbf{53.6} $ & $ \mathbf{53.5} $ \\
    \cmidrule(r){1-1} \cmidrule(lr){2-2} \cmidrule(l){3-5}
    DBS+$k$NN-MT & $ 42.0 $ & $ 48.6 $ & - & $ 51.8 $ \\
    \quad+Static & $ 55.2 $ & $ 49.0 $ & - & $ 52.0 $ \\
    \quad+Adaptive & $ 53.7 $ & $ 49.0 $ & - & $ 52.1 $ \\
    \quad+Randomize & $ 54.4 $ & $ 48.4 $ & - & $ 51.5 $ \\
    Nucleus+$k$NN-MT & $ 51.6 $ & $ 50.4 $ & - & $ 52.8 $ \\
    \quad+Static & $ 55.0 $ & $ 49.9 $ & - & $ 52.5 $ \\
    \quad+Adaptive & $ 55.6 $ & $ 49.8 $ & - & $ 52.4 $ \\
    \quad+Randomize & $ \mathbf{59.4} $ & $ 49.2 $ & - & $ 52.0 $ \\
    \bottomrule
  \end{tabular}
  \caption{Ablation study for MergedBLEU@40 in the De-En domain adaptation setting: DP, BLEU@20 and MergedBLEU@40 are the scores when beam size is set to 20, and BLEU@40 is the score when beam size is set to 40. We report averages of five domains.}\label{tab:oracle40_de-en}
\end{table}

\begin{table}[tbp]
  \centering
  \small
  \begin{tabular}{@{}lcccc@{}}
    \toprule
    & \multirow{2}{*}{DP} & \multicolumn{3}{c}{BLEU} \\
    Method & & @20 & @40 & Mrg@40 \\
    \cmidrule[.05em](r){1-1} \cmidrule[.05em](lr){2-2} \cmidrule[.05em](l){3-5}
    Baseline & $ 38.0 $ & $ 26.0 $ & $ 28.2 $ & - \\
    DBS & $ 54.9 $ & $ 24.8 $ & $ 26.9 $ & $ 28.2 $ \\
    Nucleus & $ 63.9 $ & $ 26.5 $ & $ 28.4 $ & $ 28.9 $ \\
    $k$NN-MT & $ 37.4 $ & $ \mathbf{29.7} $ & $ \mathbf{31.8} $ & $ 31.5 $ \\
    \cmidrule(r){1-1} \cmidrule(lr){2-2} \cmidrule(l){3-5}
    DBS+$k$NN-MT & $ 60.7 $ & $ 27.9 $ & - & $ 30.9 $ \\
    \quad+Static & $ 66.5 $ & $ 28.3 $ & - & $ 31.3 $ \\
    \quad+Adaptive & $ 66.8 $ & $ 28.4 $ & - & $ 31.3 $ \\
    \quad+Randomize & $ 65.9 $ & $ 27.8 $ & - & $ 30.9 $ \\
    Nucleus+$k$NN-MT & $ 66.6 $ & $ 29.0 $ & - & $ \mathbf{31.7} $ \\
    \quad+Static & $ 64.0 $ & $ 28.9 $ & - & $ 31.5 $ \\
    \quad+Adaptive & $ 64.0 $ & $ 28.8 $ & - & $ 31.5 $ \\
    \quad+Randomize & $ \mathbf{74.8} $ & $ 28.7 $ & - & $ 31.4 $ \\
    \bottomrule
  \end{tabular}
  \caption{Ablation study for MergedBLEU@40 in the Ja-En domain adaptation setting: DP, BLEU@20 and MergedBLEU@40 are the scores when beam size is set to 20, and BLEU@40 is the score when beam size is set to 40. We report averages of four domains.}\label{tab:oracle40_ja-en}
\end{table}

\begin{table}[tbp]
  \centering
  \small
  \begin{tabular}{@{}lcccc@{}}
    \toprule
    & \multirow{2}{*}{DP} & \multicolumn{3}{c}{BLEU} \\
    Method & & @20 & @40 & Mrg@40 \\
    \cmidrule[.05em](r){1-1} \cmidrule[.05em](lr){2-2} \cmidrule[.05em](l){3-5}
    Baseline & $ 37.5 $ & $ 41.2 $ & $ 43.9 $ & - \\
    DBS & $ 51.3 $ & $ 37.9 $ & $ 40.2 $ & $ 43.1 $ \\
    Nucleus & $ 62.8 $ & $ 40.4 $ & $ 43.1 $ & $ 44.2 $ \\
    $k$NN-MT & $ 37.3 $ & $ \mathbf{41.4} $ & $ \mathbf{44.0} $ & $ 42.6 $ \\
    \cmidrule(r){1-1} \cmidrule(lr){2-2} \cmidrule(l){3-5}
    DBS+$k$NN-MT & $ 52.6 $ & $ 38.2 $ & - & $ 43.4 $ \\
    \quad+Static & $ 54.8 $ & $ 38.3 $ & - & $ 43.4 $ \\
    \quad+Adaptive & $ 54.3 $ & $ 38.3 $ & - & $ 43.4 $ \\
    \quad+Randomize & $ 53.9 $ & $ 38.2 $ & - & $ 43.4 $ \\
    \quad+Uniquify & $ 54.9 $ & $ 37.8 $ & - & $ 43.2 $ \\
    \quad\quad+Static & $ 55.9 $ & $ 37.7 $ & - & $ 43.1 $ \\
    \quad\quad+Adaptive & $ 55.7 $ & $ 37.8 $ & - & $ 43.2 $ \\
    \quad\quad+Randomize & $ 55.8 $ & $ 37.7 $ & - & $ 43.2 $ \\
    \cmidrule(r){1-1} \cmidrule(lr){2-2} \cmidrule(l){3-5}
    Nucleus+$k$NN-MT & $ 64.5 $ & $ 40.5 $ & - & $ \mathbf{44.3} $ \\
    \quad+Static & $ 52.4 $ & $ 38.9 $ & - & $ 43.4 $ \\
    \quad+Adaptive & $ 52.8 $ & $ 38.9 $ & - & $ 43.4 $ \\
    \quad+Randomize & $ 62.3 $ & $ 38.9 $ & - & $ 43.5 $ \\
    \quad+Uniquify & $ \mathbf{70.8} $ & $ 39.3 $ & - & $ 44.0 $ \\
    \quad\quad+Static & $ 55.4 $ & $ 38.7 $ & - & $ 43.3 $ \\
    \quad\quad+Adaptive & $ 55.7 $ & $ 38.7 $ & - & $ 43.4 $ \\
    \quad\quad+Randomize & $ 67.7 $ & $ 38.4 $ & - & $ 43.3 $ \\
    \bottomrule
  \end{tabular}
  \caption{Ablation study for MergedBLEU@40 in the general-domain setting: DP, BLEU@20 and MergedBLEU@40 are the scores when beam size is set to 20, and BLEU@40 is the score when beam size is set to 40. We report averages of three language pairs.}\label{tab:oracle40_general}
\end{table}

\subsection{\texorpdfstring{Tuning $k$NN Diversified Decoding}{}}
\label{appendix:tuning_proposed_method}
\begin{figure*}[tbp]
    \centering
    \begin{tabular}{cc}
    \begin{subfigure}[t]{0.45\textwidth}
      \centering
      \includegraphics[width=\textwidth]{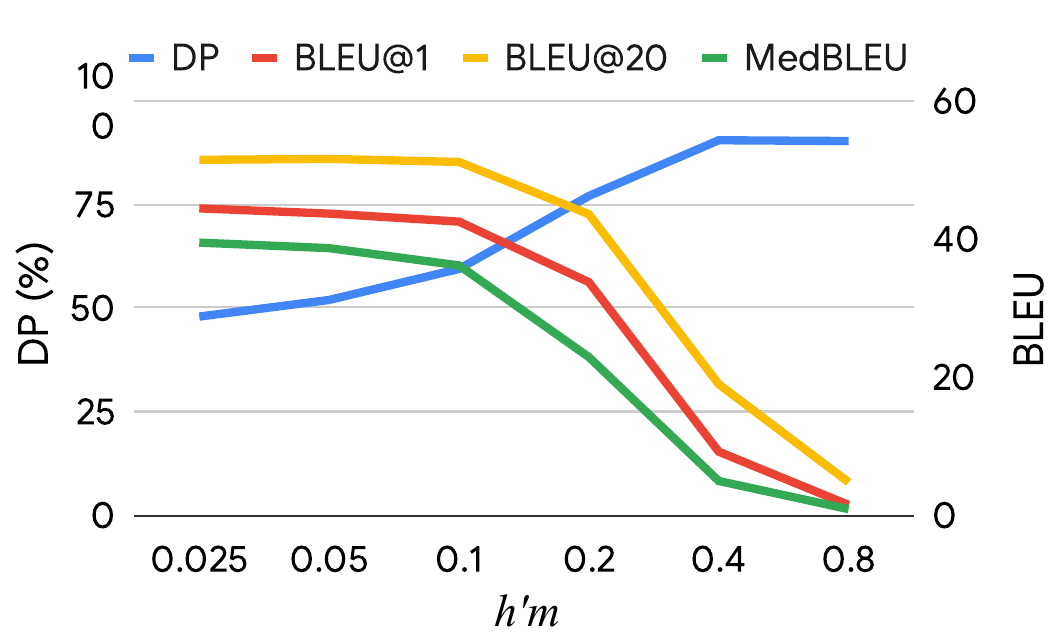}
      \caption{Adaptive noise}
      \label{fig:tradeoff_adaptive}
    \end{subfigure}
    &
    \begin{subfigure}[t]{0.45\textwidth}
      \centering
      \includegraphics[width=\textwidth]{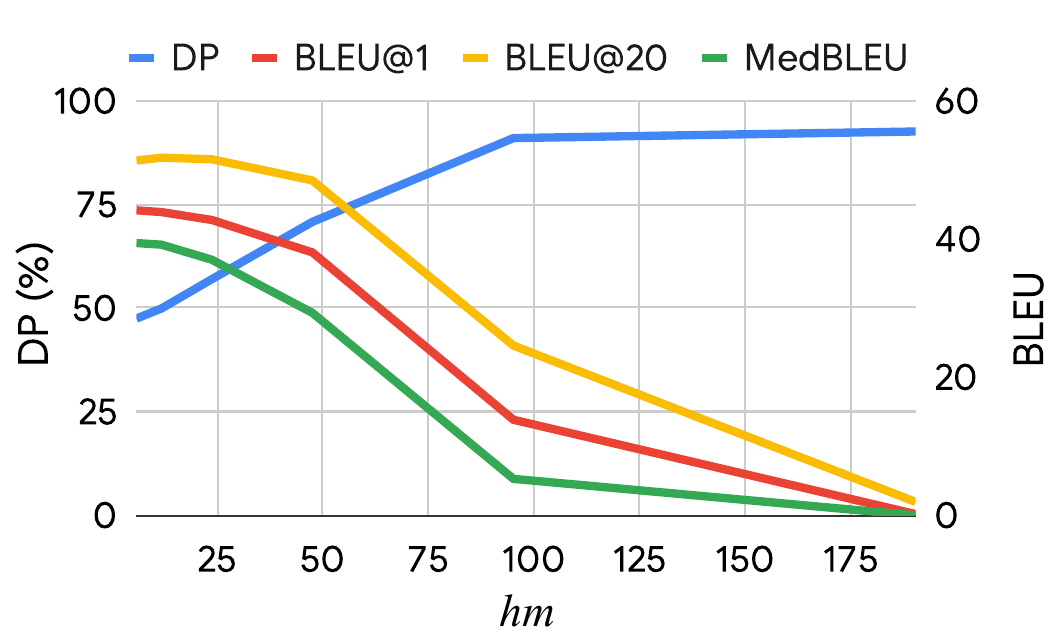}
      \caption{Static noise}
      \label{fig:tradeoff_static}
    \end{subfigure}
    \\
    \begin{subfigure}[t]{0.45\textwidth}
      \centering
      \includegraphics[width=\textwidth]{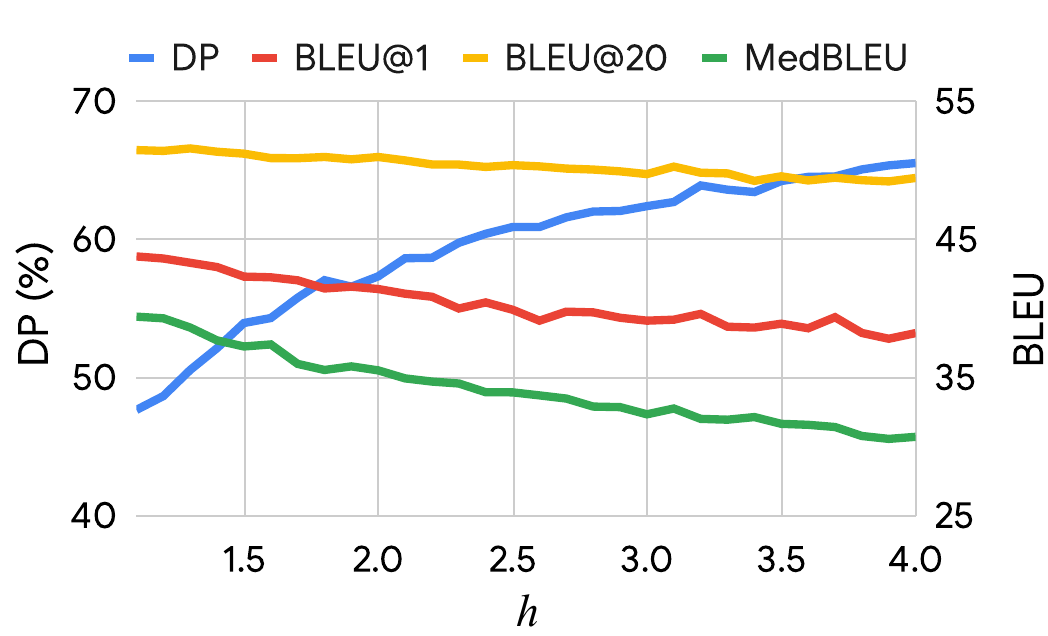}
      \caption{Randomized-$k$NN}
      \label{fig:tradeoff_randomized-k}
    \end{subfigure}
    &
    \begin{subfigure}[t]{0.45\textwidth}
      \centering
      \includegraphics[width=\textwidth]{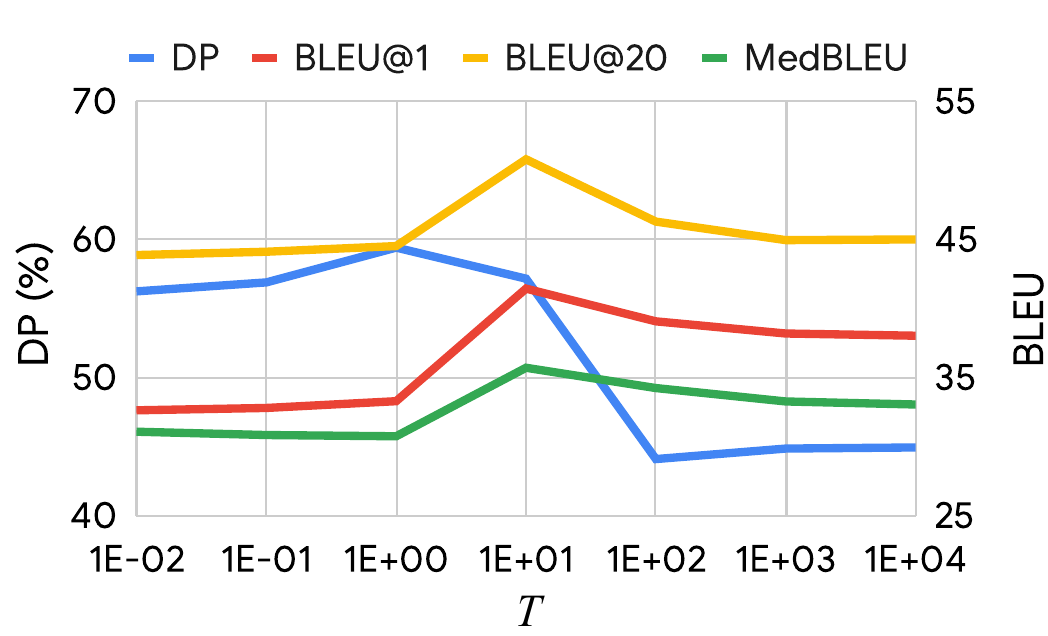}
      \caption{Temperature}
      \label{fig:tradeoff_temperature}
    \end{subfigure}
    \end{tabular}\\
    \caption{Relationship among perturbation's magnitudes or temperature and DP/BLEU on the De-En IT domain}
    \label{fig:tradeoff_appendix}
\end{figure*}

Figure~\ref{fig:tradeoff_appendix} shows the relationship of the magnitude of the perturbation against DP and BLEU of DBS+$k$NN-MT+Perturbation in the De-En IT domain.
This result shows the trade-off between the DP and the BLEUs for all the perturbation types, indicating that the proposed methods adjust the diversity and the translation quality by varying the perturbation's magnitude. 
The effect of temperature $\tau$ on $k$NN probability of DBS+$k$NN-MT is also shown in Figure~\ref{fig:tradeoff_temperature}.
Both DP and BLEU peak around $\tau$ from 1 to 10.
Unlike the perturbation's magnitude, we found no trade-off between DP and BLEU for the temperature adjustment.

\subsection{Detailed Quantitative Analysis}
\label{sec:appendix_quantitative}
Figure~\ref{fig:ex_full} shows a detailed quantitative analysis.

\section{Used Data, Model, and Software}
\subsection{Data}
\begin{description}
\item[De-En domain adaptation parallel corpora] created by \newcite{koehn-knowles-2017-six} based on OPUS~\cite{tiedemann-2012-parallel}. License: allowed for research purpose use.
\item[The Asian Scientific Paper Excerpt Corpus] created by \newcite{nakazawa-etal-2016-aspec}. License: \url{https://jipsti.jst.go.jp/aspec/}.
\item[The Kyoto Free Translation Task] created by \newcite{neubig11kftt}. Download: \url{http://www.phontron.com/kftt/index.html}, License: CC BY-SA 3.0.
\item[Ted talks] created by \newcite{cettolo-etal-2012-wit3}. Download: \url{https://wit3.fbk.eu/}, License: CC BY-NC-ND.
\item[The Business Scene Dialogue corpus] created by \newcite{rikters-etal-2019-designing}, Download: \url{https://github.com/tsuruoka-lab/BSD}, License: CC BY-NC-SA.
\item[WMT'19 news translation task] created by \newcite{barrault-etal-2019-findings}, Download: \url{https://www.statmt.org/wmt19/translation-task.html}, License: allowed for research purpose use.
\item[WMT'21 news translation task] created by \newcite{akhbardeh-etal-2021-findings}, Download: \url{https://www.statmt.org/wmt21/translation-task.html}, License: allowed for research purpose use.
\item[WMT'22 general translation task] created by \newcite{kocmi-EtAl:2022:WMT}, Download: \url{https://www.statmt.org/wmt22/translation-task.html}, License: allowed for research purpose use.
\item[JParaCrawl v3.0] created by \newcite{morishita-etal-2022-jparacrawl}. Download: \url{http://www.kecl.ntt.co.jp/icl/lirg/jparacrawl/}, License: allowed for research purpose use.
\end{description}

\subsection{Model}
\begin{description}
\item[WMT'19 De-En pre-trained model] trained by \newcite{ng-etal-2019-facebook}. Download: \url{https://github.com/facebookresearch/fairseq/tree/main/examples/wmt19}, License: MIT.
\end{description}

\subsection{Software}
\begin{description}
\item[\texttt{fairseq}] created by \newcite{ott-etal-2019-fairseq}. Download: \url{https://github.com/facebookresearch/fairseq}, License: MIT.
\item[\texttt{FAISS}] created by \newcite{johnson2019billion}. Download: \url{https://github.com/facebookresearch/faiss}, License: MIT.
\item[\texttt{sacreBLEU}] created by \newcite{post-2018-call}. Download: \url{https://github.com/mjpost/sacrebleu}, License: Apache License 2.0. 
\item[\texttt{COMET}] created by \newcite{rei-etal-2020-comet}. Download: \url{https://github.com/Unbabel/COMET}, License: Apache License 2.0. 
\item[\texttt{BERTScore}] created by \newcite{Zhang*2020BERTScore:}. Download: \url{https://github.com/Tiiiger/bert_score}, License: MIT. 
\end{description}

\begin{table*}[htbp]
  \centering
  \footnotesize
  \tabcolsep 4pt
\begin{tabular}{c|ccccc|cccc|cccc}
    \toprule
    & \multicolumn{5}{c|}{De-En Domain Adaptation} & \multicolumn{4}{c|}{Ja-En Domain Adaptation} & \multicolumn{3}{c}{General Domain}\\
    Hyperparameters & Koran & IT & Medical & Law & Subtitles & ASPEC & KFTT & BSD & TED & De-En & Ja-En & Uk-Cs \\ \midrule
    \multicolumn{13}{c}{Nucleus}  \\ \cmidrule{1-13}
    $p$ & 0.6 & 0.8 & 0.7 & 0.7 & 0.8 & 0.6 & 0.6 & 0.6 & 0.5 & 0.7 & 0.6 & 0.6 \\ \midrule
    \multicolumn{13}{c}{$k$NN-MT} \\ \cmidrule{1-13}
    $k$ & 64 & 64 & 64 & 64 & 64 & 64 & 64 & 64 & 64 & 32 & 16 & 16 \\
    $\tau$ & 100 & 10 & 10 & 10 & 10 & 100 & 100 & 10 & 100 & 100 & 100 & 100 \\
    $\lambda$ & 0.8 & 0.7 & 0.8 & 0.8 & 0.7 & 0.7 & 0.6 & 0.3 & 0.6 & 0.2 & 0.2 & 0.2 \\
\bottomrule
  \end{tabular}
  \caption{Hyperparameters}
\label{tab:hyperparameter}
\end{table*}

\begin{table*}[htbp]
  \centering
  \small
  \tabcolsep 2.3pt
  \begin{tabular}{@{}lcccccccccccccccc@{}}
    \toprule
    & \multirow{2}{*}{DP} & \multicolumn{4}{c}{BLEU} & \multirow{2}{*}{DEQ} & \multirow{2}{*}{MLen} & \multicolumn{2}{c}{COMET} & \multicolumn{2}{c}{BERTScore} & \multicolumn{3}{c}{PLL} & \multirow{2}{*}{Speed}\\
    Method & & @1 & @20 & Mrg & Ref & & & @1 & @20 & @1 & @20 & Max & Min & Mean & \\
    \cmidrule[.05em](r){1-1} \cmidrule[.05em](lr){2-2} \cmidrule[.05em](lr){3-6} \cmidrule[.05em](lr){7-7} \cmidrule[.05em](lr){8-8} \cmidrule[.05em](lr){9-10} \cmidrule[.05em](lr){11-12} \cmidrule[.05em](lr){13-15} \cmidrule[.05em](l){16-16}
    Reference & - & - & - & - & - & - & - & - & - & - & - & - & - & $ -2.95 $ & - \\
    \cmidrule(r){1-1} \cmidrule(lr){2-2} \cmidrule(lr){3-6} \cmidrule(lr){7-7} \cmidrule(lr){8-8} \cmidrule(lr){9-10} \cmidrule(lr){11-12} \cmidrule(lr){13-15} \cmidrule(l){16-16}
    Baseline & $ 27.4 $ & $ 16.9 $ & $ 22.5 $ & - & $ 16.3 $ & $ 0.33 $ & $ 0.950 $ & $ 0.723 $ & $ 0.763 $ & $ 0.389 $ & $ 0.464 $ & $ -2.08 $ & $ -3.48 $ & $ -2.73 $ & $ 897.2 $ \\
    DBS & $ 39.4 $ & $ 17.0 $ & $ 22.4 $ & $ 24.3 $ & $ 15.5 $ & $ 2.94 $ & $ 0.937 $ & $ 0.720 $ & $ 0.763 $ & $ 0.392 $ & $ 0.465 $ & $ -1.97 $ & $ -3.74 $ & $ -2.77 $ & $ 565.2 $ \\
    Nucleus & $ 52.4 $ & $ 16.7 $ & $ 23.8 $ & $ 25.1 $ & $ 15.5 $ & $ 5.80 $ & $ 0.955 $ & $ 0.722 $ & $ 0.766 $ & $ 0.387 $ & $ 0.466 $ & $ -2.03 $ & $ -3.57 $ & $ -2.76 $ & $ 519.4 $ \\
    $k$NN-MT & $ 26.2 $ & $ 21.0 $ & $ 27.4 $ & $ 29.0 $ & $ 20.0 $ & - & $ 0.946 $ & $ 0.728 $ & $ 0.775 $ & $ 0.423 $ & $ 0.514 $ & $ -2.01 $ & $ -3.48 $ & $ -2.69 $ & $ 86.8 $ \\
    \cmidrule(r){1-1} \cmidrule(lr){2-2} \cmidrule(lr){3-6} \cmidrule(lr){7-7} \cmidrule(lr){8-8} \cmidrule(lr){9-10} \cmidrule(lr){11-12} \cmidrule(lr){13-15} \cmidrule(l){16-16}
    DBS+$k$NN-MT & $ 47.2 $ & $ 20.5 $ & $ 27.0 $ & $ 28.8 $ & $ 17.9 $ & $ 10.22 $ & $ 0.945 $ & $ 0.723 $ & $ 0.775 $ & $ 0.424 $ & $ 0.511 $ & $ -1.72 $ & $ -3.84 $ & $ -2.73 $ & $ 75.8 $ \\
\bottomrule
  \end{tabular}
  \caption{Koran domain in German-English}\label{tab:koran}
\end{table*}

\begin{table*}[htbp]
  \centering
  \small
  \tabcolsep 2.3pt
  \begin{tabular}{@{}lcccccccccccccccc@{}}
  \toprule
    & \multirow{2}{*}{DP} & \multicolumn{4}{c}{BLEU} & \multirow{2}{*}{DEQ} & \multirow{2}{*}{MLen} & \multicolumn{2}{c}{COMET} & \multicolumn{2}{c}{BERTScore} & \multicolumn{3}{c}{PLL} & \multirow{2}{*}{Speed}\\
    Method & & @1 & @20 & Mrg & Ref & & & @1 & @20 & @1 & @20 & Max & Min & Mean & \\
    \cmidrule[.05em](r){1-1} \cmidrule[.05em](lr){2-2} \cmidrule[.05em](lr){3-6} \cmidrule[.05em](lr){7-7} \cmidrule[.05em](lr){8-8} \cmidrule[.05em](lr){9-10} \cmidrule[.05em](lr){11-12} \cmidrule[.05em](lr){13-15} \cmidrule[.05em](l){16-16}
    Reference & - & - & - & - & - & - & - & - & - & - & - & - & - & $ -4.93 $ & - \\
    \cmidrule(r){1-1} \cmidrule(lr){2-2} \cmidrule(lr){3-6} \cmidrule(lr){7-7} \cmidrule(lr){8-8} \cmidrule(lr){9-10} \cmidrule(lr){11-12} \cmidrule(lr){13-15} \cmidrule(l){16-16}
    Baseline & $ 31.5 $ & $ 37.7 $ & $ 47.3 $ & - & $ 34.0 $ & $ -0.21 $ & $ 1.004 $ & $ 0.821 $ & $ 0.873 $ & $ 0.606 $ & $ 0.729 $ & $ -2.96 $ & $ -7.21 $ & $ -4.87 $ & $ 874.8 $ \\
    DBS & $ 35.2 $ & $ 37.1 $ & $ 44.2 $ & $ 48.5 $ & $ 33.9 $ & $ 0.41 $ & $ 0.998 $ & $ 0.821 $ & $ 0.861 $ & $ 0.606 $ & $ 0.696 $ & $ -2.97 $ & $ -7.05 $ & $ -4.82 $ & $ 597.7 $ \\
    Nucleus & $ 51.8 $ & $ 36.4 $ & $ 45.5 $ & $ 48.8 $ & $ 31.9 $ & $ 2.42 $ & $ 1.011 $ & $ 0.807 $ & $ 0.856 $ & $ 0.574 $ & $ 0.684 $ & $ -3.09 $ & $ -6.97 $ & $ -4.90 $ & $ 487.1 $ \\
    $k$NN-MT & $ 32.7 $ & $ 45.9 $ & $ 55.0 $ & $ 57.0 $ & $ 39.8 $ & - & $ 0.974 $ & $ 0.829 $ & $ 0.891 $ & $ 0.645 $ & $ 0.782 $ & $ -2.94 $ & $ -7.69 $ & $ -5.04 $ & $ 59.5 $ \\
    \cmidrule(r){1-1} \cmidrule(lr){2-2} \cmidrule(lr){3-6} \cmidrule(lr){7-7} \cmidrule(lr){8-8} \cmidrule(lr){9-10} \cmidrule(lr){11-12} \cmidrule(lr){13-15} \cmidrule(l){16-16}
    DBS+$k$NN-MT & $ 44.6 $ & $ 43.9 $ & $ 50.9 $ & $ 55.4 $ & $ 37.4 $ & $ 4.94 $ & $ 0.975 $ & $ 0.815 $ & $ 0.874 $ & $ 0.617 $ & $ 0.735 $ & $ -3.09 $ & $ -7.60 $ & $ -5.05 $ & $ 57.3 $ \\
\bottomrule
  \end{tabular}
  \caption{IT domain in German-English}\label{tab:it}
\end{table*}

\begin{table*}[htbp]
    \centering
    \small
    \tabcolsep 2.3pt
    \begin{tabular}{@{}lcccccccccccccccc@{}}
    \toprule
    & \multirow{2}{*}{DP} & \multicolumn{4}{c}{BLEU} & \multirow{2}{*}{DEQ} & \multirow{2}{*}{MLen} & \multicolumn{2}{c}{COMET} & \multicolumn{2}{c}{BERTScore} & \multicolumn{3}{c}{PLL} & \multirow{2}{*}{Speed}\\
    Method & & @1 & @20 & Mrg & Ref & & & @1 & @20 & @1 & @20 & Max & Min & Mean & \\
    \cmidrule[.05em](r){1-1} \cmidrule[.05em](lr){2-2} \cmidrule[.05em](lr){3-6} \cmidrule[.05em](lr){7-7} \cmidrule[.05em](lr){8-8} \cmidrule[.05em](lr){9-10} \cmidrule[.05em](lr){11-12} \cmidrule[.05em](lr){13-15} \cmidrule[.05em](l){16-16}  
    Reference & - & - & - & - & - & - & - & - & - & - & - & - & - & $ -3.24 $ & - \\
    \cmidrule(r){1-1} \cmidrule(lr){2-2} \cmidrule(lr){3-6} \cmidrule(lr){7-7} \cmidrule(lr){8-8} \cmidrule(lr){9-10} \cmidrule(lr){11-12} \cmidrule(lr){13-15} \cmidrule(l){16-16}
    Baseline & $ 27.7 $ & $ 40.4 $ & $ 49.1 $ & - & $ 37.3 $ & $ -0.15 $ & $ 0.946 $ & $ 0.832 $ & $ 0.858 $ & $ 0.643 $ & $ 0.711 $ & $ -2.33 $ & $ -4.23 $ & $ -3.17 $ & $ 706.0 $ \\
    DBS & $ 31.4 $ & $ 39.9 $ & $ 46.0 $ & $ 50.2 $ & $ 36.3 $ & $ 0.16 $ & $ 0.938 $ & $ 0.831 $ & $ 0.854 $ & $ 0.641 $ & $ 0.697 $ & $ -2.26 $ & $ -4.29 $ & $ -3.15 $ & $ 401.2 $ \\
    Nucleus & $ 39.2 $ & $ 40.0 $ & $ 48.6 $ & $ 50.9 $ & $ 37.1 $ & $ 0.82 $ & $ 0.950 $ & $ 0.825 $ & $ 0.853 $ & $ 0.630 $ & $ 0.694 $ & $ -2.42 $ & $ -4.10 $ & $ -3.20 $ & $ 479.4 $ \\
    $k$NN-MT & $ 29.5 $ & $ 55.4 $ & $ 63.0 $ & $ 64.4 $ & $ 48.8 $ & $ 0.00 $ & $ 0.928 $ & $ 0.847 $ & $ 0.875 $ & $ 0.707 $ & $ 0.776 $ & $ -2.31 $ & $ -4.59 $ & $ -3.28 $ & $ 17.9 $ \\
    \cmidrule(r){1-1} \cmidrule(lr){2-2} \cmidrule(lr){3-6} \cmidrule(lr){7-7} \cmidrule(lr){8-8} \cmidrule(lr){9-10} \cmidrule(lr){11-12} \cmidrule(lr){13-15} \cmidrule(l){16-16}
    DBS+$k$NN-MT & $ 36.7 $ & $ 54.0 $ & $ 59.6 $ & $ 62.3 $ & $ 47.1 $ & $ 4.22 $ & $ 0.937 $ & $ 0.836 $ & $ 0.868 $ & $ 0.684 $ & $ 0.753 $ & $ -2.31 $ & $ -4.77 $ & $ -3.31 $ & $ 15.7 $ \\
\bottomrule
    \end{tabular}
    \caption{Medical domain in German-English}\label{tab:medical}
\end{table*}

\begin{table*}[htbp]
    \centering
    \small
    \tabcolsep 2.3pt
    \begin{tabular}{@{}lcccccccccccccccc@{}}
    \toprule
    & \multirow{2}{*}{DP} & \multicolumn{4}{c}{BLEU} & \multirow{2}{*}{DEQ} & \multirow{2}{*}{MLen} & \multicolumn{2}{c}{COMET} & \multicolumn{2}{c}{BERTScore} & \multicolumn{3}{c}{PLL} & \multirow{2}{*}{Speed}\\
    Method & & @1 & @20 & Mrg & Ref & & & @1 & @20 & @1 & @20 & Max & Min & Mean & \\
    \cmidrule[.05em](r){1-1} \cmidrule[.05em](lr){2-2} \cmidrule[.05em](lr){3-6} \cmidrule[.05em](lr){7-7} \cmidrule[.05em](lr){8-8} \cmidrule[.05em](lr){9-10} \cmidrule[.05em](lr){11-12} \cmidrule[.05em](lr){13-15} \cmidrule[.05em](l){16-16}
    Reference & - & - & - & - & - & - & - & - & - & - & - & - & - & $ -2.57 $ & - \\
    \cmidrule(r){1-1} \cmidrule(lr){2-2} \cmidrule(lr){3-6} \cmidrule(lr){7-7} \cmidrule(lr){8-8} \cmidrule(lr){9-10} \cmidrule(lr){11-12} \cmidrule(lr){13-15} \cmidrule(l){16-16}
    Baseline & $ 19.5 $ & $ 46.1 $ & $ 52.4 $ & - & $ 44.1 $ & $ -0.01 $ & $ 0.963 $ & $ 0.854 $ & $ 0.873 $ & $ 0.675 $ & $ 0.731 $ & $ -1.97 $ & $ -3.17 $ & $ -2.50 $ & $ 822.4 $ \\
    DBS & $ 27.4 $ & $ 45.0 $ & $ 50.2 $ & $ 53.6 $ & $ 41.7 $ & $ 0.46 $ & $ 0.936 $ & $ 0.851 $ & $ 0.869 $ & $ 0.670 $ & $ 0.717 $ & $ -2.00 $ & $ -3.35 $ & $ -2.58 $ & $ 428.1 $ \\
    Nucleus & $ 43.2 $ & $ 44.7 $ & $ 52.9 $ & $ 54.8 $ & $ 41.2 $ & $ 1.35 $ & $ 0.957 $ & $ 0.850 $ & $ 0.873 $ & $ 0.665 $ & $ 0.730 $ & $ -1.94 $ & $ -3.22 $ & $ -2.54 $ & $ 458.4 $ \\
    $k$NN-MT & $ 19.6 $ & $ 61.9 $ & $ 68.8 $ & $ 69.9 $ & $ 58.6 $ & - & $ 0.977 $ & $ 0.871 $ & $ 0.891 $ & $ 0.757 $ & $ 0.818 $ & $ -2.02 $ & $ -3.27 $ & $ -2.57 $ & $ 27.2 $ \\
    \cmidrule(r){1-1} \cmidrule(lr){2-2} \cmidrule(lr){3-6} \cmidrule(lr){7-7} \cmidrule(lr){8-8} \cmidrule(lr){9-10} \cmidrule(lr){11-12} \cmidrule(lr){13-15} \cmidrule(l){16-16}
    DBS+$k$NN-MT & $ 26.9 $ & $ 60.8 $ & $ 65.8 $ & $ 67.5 $ & $ 55.9 $ & $ 2.69 $ & $ 0.965 $ & $ 0.863 $ & $ 0.885 $ & $ 0.743 $ & $ 0.798 $ & $ -2.06 $ & $ -3.44 $ & $ -2.64 $ & $ 26.1 $ \\
\bottomrule
    \end{tabular}
    \caption{Law domain in German-English}\label{tab:law}
\end{table*}

\begin{table*}[htbp]
  \centering
  \small
  \tabcolsep 2.3pt
  \begin{tabular}{@{}lcccccccccccccccc@{}}
    \toprule
    & \multirow{2}{*}{DP} & \multicolumn{4}{c}{BLEU} & \multirow{2}{*}{DEQ} & \multirow{2}{*}{MLen} & \multicolumn{2}{c}{COMET} & \multicolumn{2}{c}{BERTScore} & \multicolumn{3}{c}{PLL} & \multirow{2}{*}{Speed}\\
    Method & & @1 & @20 & Mrg & Ref & & & @1 & @20 & @1 & @20 & Max & Min & Mean & \\
    \cmidrule[.05em](r){1-1} \cmidrule[.05em](lr){2-2} \cmidrule[.05em](lr){3-6} \cmidrule[.05em](lr){7-7} \cmidrule[.05em](lr){8-8} \cmidrule[.05em](lr){9-10} \cmidrule[.05em](lr){11-12} \cmidrule[.05em](lr){13-15} \cmidrule[.05em](l){16-16}
    Reference & - & - & - & - & - & - & - & - & - & - & - & - & - & $ -3.07 $ & - \\
    \cmidrule(r){1-1} \cmidrule(lr){2-2} \cmidrule(lr){3-6} \cmidrule(lr){7-7} \cmidrule(lr){8-8} \cmidrule(lr){9-10} \cmidrule(lr){11-12} \cmidrule(lr){13-15} \cmidrule(l){16-16}
    Baseline & $ 51.0 $ & $ 29.3 $ & $ 41.7 $ & - & $ 23.0 $ & $ -1.66 $ & $ 0.917 $ & $ 0.799 $ & $ 0.843 $ & $ 0.595 $ & $ 0.685 $ & $ -1.98 $ & $ -4.65 $ & $ -3.12 $ & $ 1095.9 $ \\
    DBS & $ 45.9 $ & $ 29.0 $ & $ 36.9 $ & $ 42.6 $ & $ 24.1 $ & $ -14.20 $ & $ 0.923 $ & $ 0.797 $ & $ 0.833 $ & $ 0.593 $ & $ 0.660 $ & $ -1.96 $ & $ -4.74 $ & $ -3.10 $ & $ 857.7 $ \\
    Nucleus & $ 53.2 $ & $ 29.0 $ & $ 39.8 $ & $ 43.3 $ & $ 24.6 $ & $ -14.33 $ & $ 0.935 $ & $ 0.798 $ & $ 0.837 $ & $ 0.593 $ & $ 0.668 $ & $ -2.07 $ & $ -4.25 $ & $ -3.04 $ & $ 700.4 $ \\
    $k$NN-MT & $ 53.6 $ & $ 31.7 $ & $ 45.1 $ & $ 47.1 $ & $ 24.6 $ & - & $ 0.911 $ & $ 0.801 $ & $ 0.851 $ & $ 0.606 $ & $ 0.709 $ & $ -1.87 $ & $ -4.65 $ & $ -3.04 $ & $ 4.5 $ \\
    \cmidrule(r){1-1} \cmidrule(lr){2-2} \cmidrule(lr){3-6} \cmidrule(lr){7-7} \cmidrule(lr){8-8} \cmidrule(lr){9-10} \cmidrule(lr){11-12} \cmidrule(lr){13-15} \cmidrule(l){16-16}
    DBS+$k$NN-MT & $ 54.6 $ & $ 30.6 $ & $ 39.5 $ & $ 45.0 $ & $ 24.4 $ & $ 3.45 $ & $ 0.930 $ & $ 0.793 $ & $ 0.839 $ & $ 0.597 $ & $ 0.677 $ & $ -1.70 $ & $ -4.86 $ & $ -3.02 $ & $ 4.4 $ \\
\bottomrule
  \end{tabular}
  \caption{Subtitles domain in German-English}\label{tab:subtitles}
\end{table*}

\begin{table*}[htbp]
  \centering
  \small
  \tabcolsep 2.3pt
  \begin{tabular}{@{}lcccccccccccccccc@{}}
    \toprule
    & \multirow{2}{*}{DP} & \multicolumn{4}{c}{BLEU} & \multirow{2}{*}{DEQ} & \multirow{2}{*}{MLen} & \multicolumn{2}{c}{COMET} & \multicolumn{2}{c}{BERTScore} & \multicolumn{3}{c}{PLL} & \multirow{2}{*}{Speed}\\
    Method & & @1 & @20 & Mrg & Ref & & & @1 & @20 & @1 & @20 & Max & Min & Mean & \\
    \cmidrule[.05em](r){1-1} \cmidrule[.05em](lr){2-2} \cmidrule[.05em](lr){3-6} \cmidrule[.05em](lr){7-7} \cmidrule[.05em](lr){8-8} \cmidrule[.05em](lr){9-10} \cmidrule[.05em](lr){11-12} \cmidrule[.05em](lr){13-15} \cmidrule[.05em](l){16-16}
    Reference & - & - & - & - & - & - & - & - & - & - & - & - & - & $ -2.90 $ & - \\
    \cmidrule(r){1-1} \cmidrule(lr){2-2} \cmidrule(lr){3-6} \cmidrule(lr){7-7} \cmidrule(lr){8-8} \cmidrule(lr){9-10} \cmidrule(lr){11-12} \cmidrule(lr){13-15} \cmidrule(l){16-16}
    Baseline & $ 31.2 $ & $ 21.0 $ & $ 28.2 $ & - & $ 19.8 $ & $ 0.07 $ & $ 0.946 $ & $ 0.829 $ & $ 0.849 $ & $ 0.572 $ & $ 0.629 $ & $ -1.86 $ & $ -2.83 $ & $ -2.31 $ & $ 914.3 $ \\
    DBS & $ 50.1 $ & $ 19.5 $ & $ 27.4 $ & $ 30.6 $ & $ 17.0 $ & $ 3.12 $ & $ 0.904 $ & $ 0.826 $ & $ 0.850 $ & $ 0.563 $ & $ 0.629 $ & $ -1.76 $ & $ -3.24 $ & $ -2.43 $ & $ 581.3 $ \\
    Nucleus & $ 64.9 $ & $ 20.4 $ & $ 29.7 $ & $ 31.7 $ & $ 16.9 $ & $ 5.40 $ & $ 0.934 $ & $ 0.827 $ & $ 0.853 $ & $ 0.566 $ & $ 0.637 $ & $ -1.66 $ & $ -3.27 $ & $ -2.40 $ & $ 591.7 $ \\
    $k$NN-MT & $ 31.0 $ & $ 24.7 $ & $ 32.8 $ & $ 34.6 $ & $ 23.1 $ & - & $ 0.955 $ & $ 0.831 $ & $ 0.853 $ & $ 0.594 $ & $ 0.658 $ & $ -1.86 $ & $ -2.91 $ & $ -2.33 $ & $ 8.2 $ \\
    \cmidrule(r){1-1} \cmidrule(lr){2-2} \cmidrule(lr){3-6} \cmidrule(lr){7-7} \cmidrule(lr){8-8} \cmidrule(lr){9-10} \cmidrule(lr){11-12} \cmidrule(lr){13-15} \cmidrule(l){16-16}
    DBS+$k$NN-MT & $ 58.6 $ & $ 22.9 $ & $ 31.3 $ & $ 34.0 $ & $ 18.3 $ & $ 5.67 $ & $ 0.895 $ & $ 0.827 $ & $ 0.854 $ & $ 0.583 $ & $ 0.656 $ & $ -1.56 $ & $ -3.44 $ & $ -2.43 $ & $ 7.6 $ \\
\bottomrule
  \end{tabular}
  \caption{ASPEC domain in Japanese-English}\label{tab:aspec}
\end{table*}

\begin{table*}[htbp]
  \centering
  \small
  \tabcolsep 2.3pt
  \begin{tabular}{@{}lcccccccccccccccc@{}}
    \toprule
    & \multirow{2}{*}{DP} & \multicolumn{4}{c}{BLEU} & \multirow{2}{*}{DEQ} & \multirow{2}{*}{MLen} & \multicolumn{2}{c}{COMET} & \multicolumn{2}{c}{BERTScore} & \multicolumn{3}{c}{PLL} & \multirow{2}{*}{Speed}\\
    Method & & @1 & @20 & Mrg & Ref & & & @1 & @20 & @1 & @20 & Max & Min & Mean & \\
    \cmidrule[.05em](r){1-1} \cmidrule[.05em](lr){2-2} \cmidrule[.05em](lr){3-6} \cmidrule[.05em](lr){7-7} \cmidrule[.05em](lr){8-8} \cmidrule[.05em](lr){9-10} \cmidrule[.05em](lr){11-12} \cmidrule[.05em](lr){13-15} \cmidrule[.05em](l){16-16}
    Reference & - & - & - & - & - & - & - & - & - & - & - & - & - & $ -2.75 $ & - \\
    \cmidrule(r){1-1} \cmidrule(lr){2-2} \cmidrule(lr){3-6} \cmidrule(lr){7-7} \cmidrule(lr){8-8} \cmidrule(lr){9-10} \cmidrule(lr){11-12} \cmidrule(lr){13-15} \cmidrule(l){16-16}
    Baseline & $ 31.5 $ & $ 18.8 $ & $ 25.1 $ & - & $ 17.5 $ & $ 0.70 $ & $ 0.872 $ & $ 0.762 $ & $ 0.808 $ & $ 0.490 $ & $ 0.581 $ & $ -1.88 $ & $ -3.72 $ & $ -2.70 $ & $ 758.0 $ \\
    DBS & $ 52.9 $ & $ 18.3 $ & $ 24.4 $ & $ 27.3 $ & $ 15.4 $ & $ 4.39 $ & $ 0.826 $ & $ 0.761 $ & $ 0.810 $ & $ 0.489 $ & $ 0.577 $ & $ -1.81 $ & $ -4.08 $ & $ -2.83 $ & $ 474.2 $ \\
    Nucleus & $ 68.2 $ & $ 18.6 $ & $ 26.3 $ & $ 28.2 $ & $ 15.3 $ & $ 7.07 $ & $ 0.873 $ & $ 0.764 $ & $ 0.816 $ & $ 0.493 $ & $ 0.584 $ & $ -1.85 $ & $ -3.91 $ & $ -2.82 $ & $ 444.8 $ \\
    $k$NN-MT & $ 29.2 $ & $ 22.4 $ & $ 29.4 $ & $ 31.2 $ & $ 20.8 $ & - & $ 0.913 $ & $ 0.778 $ & $ 0.822 $ & $ 0.539 $ & $ 0.628 $ & $ -1.79 $ & $ -3.75 $ & $ -2.59 $ & $ 20.5 $ \\
    \cmidrule(r){1-1} \cmidrule(lr){2-2} \cmidrule(lr){3-6} \cmidrule(lr){7-7} \cmidrule(lr){8-8} \cmidrule(lr){9-10} \cmidrule(lr){11-12} \cmidrule(lr){13-15} \cmidrule(l){16-16}
    DBS+$k$NN-MT & $ 55.3 $ & $ 21.9 $ & $ 29.0 $ & $ 30.9 $ & $ 18.1 $ & $ 9.81 $ & $ 0.870 $ & $ 0.778 $ & $ 0.827 $ & $ 0.537 $ & $ 0.627 $ & $ -1.65 $ & $ -4.28 $ & $ -2.75 $ & $ 19.4 $ \\
\bottomrule
  \end{tabular}
  \caption{KFTT domain in Japanese-English}\label{tab:kftt}
\end{table*}

\begin{table*}[htbp]
  \centering
  \small
  \tabcolsep 2.3pt
  \begin{tabular}{@{}lcccccccccccccccc@{}}
    \toprule
    & \multirow{2}{*}{DP} & \multicolumn{4}{c}{BLEU} & \multirow{2}{*}{DEQ} & \multirow{2}{*}{MLen} & \multicolumn{2}{c}{COMET} & \multicolumn{2}{c}{BERTScore} & \multicolumn{3}{c}{PLL} & \multirow{2}{*}{Speed}\\
    Method & & @1 & @20 & Mrg & Ref & & & @1 & @20 & @1 & @20 & Max & Min & Mean & \\
    \cmidrule[.05em](r){1-1} \cmidrule[.05em](lr){2-2} \cmidrule[.05em](lr){3-6} \cmidrule[.05em](lr){7-7} \cmidrule[.05em](lr){8-8} \cmidrule[.05em](lr){9-10} \cmidrule[.05em](lr){11-12} \cmidrule[.05em](lr){13-15} \cmidrule[.05em](l){16-16}
    Reference & - & - & - & - & - & - & - & - & - & - & - & - & - & $ -2.94 $ & - \\
    \cmidrule(r){1-1} \cmidrule(lr){2-2} \cmidrule(lr){3-6} \cmidrule(lr){7-7} \cmidrule(lr){8-8} \cmidrule(lr){9-10} \cmidrule(lr){11-12} \cmidrule(lr){13-15} \cmidrule(l){16-16}
    Baseline & $ 37.5 $ & $ 12.8 $ & $ 18.5 $ & - & $ 11.8 $ & $ 0.37 $ & $ 0.888 $ & $ 0.772 $ & $ 0.814 $ & $ 0.462 $ & $ 0.544 $ & $ -1.74 $ & $ -3.75 $ & $ -2.58 $ & $ 957.7 $ \\
    DBS & $ 55.6 $ & $ 12.0 $ & $ 18.0 $ & $ 20.3 $ & $ 9.9 $ & $ 4.87 $ & $ 0.839 $ & $ 0.769 $ & $ 0.815 $ & $ 0.459 $ & $ 0.541 $ & $ -1.56 $ & $ -4.01 $ & $ -2.62 $ & $ 592.9 $ \\
    Nucleus & $ 57.4 $ & $ 12.4 $ & $ 19.2 $ & $ 20.9 $ & $ 11.1 $ & $ 7.64 $ & $ 0.898 $ & $ 0.772 $ & $ 0.815 $ & $ 0.459 $ & $ 0.538 $ & $ -1.70 $ & $ -3.66 $ & $ -2.57 $ & $ 601.5 $ \\
    $k$NN-MT & $ 36.8 $ & $ 15.0 $ & $ 21.5 $ & $ 22.7 $ & $ 13.8 $ & - & $ 0.913 $ & $ 0.775 $ & $ 0.818 $ & $ 0.485 $ & $ 0.564 $ & $ -1.45 $ & $ -3.17 $ & $ -2.17 $ & $ 40.8 $ \\
    \cmidrule(r){1-1} \cmidrule(lr){2-2} \cmidrule(lr){3-6} \cmidrule(lr){7-7} \cmidrule(lr){8-8} \cmidrule(lr){9-10} \cmidrule(lr){11-12} \cmidrule(lr){13-15} \cmidrule(l){16-16}
    DBS+$k$NN-MT & $ 62.0 $ & $ 14.3 $ & $ 20.7 $ & $ 22.6 $ & $ 11.2 $ & $ 9.78 $ & $ 0.884 $ & $ 0.771 $ & $ 0.818 $ & $ 0.478 $ & $ 0.558 $ & $ -1.06 $ & $ -3.47 $ & $ -2.15 $ & $ 35.7 $ \\
\bottomrule
  \end{tabular}
  \caption{TED talks domain in Japanese-English}\label{tab:ted}
\end{table*}

\begin{table*}[htbp]
  \centering
  \small
  \tabcolsep 2.3pt
  \begin{tabular}{@{}lcccccccccccccccc@{}}
    \toprule
    & \multirow{2}{*}{DP} & \multicolumn{4}{c}{BLEU} & \multirow{2}{*}{DEQ} & \multirow{2}{*}{MLen} & \multicolumn{2}{c}{COMET} & \multicolumn{2}{c}{BERTScore} & \multicolumn{3}{c}{PLL} & \multirow{2}{*}{Speed}\\
    Method & & @1 & @20 & Mrg & Ref & & & @1 & @20 & @1 & @20 & Max & Min & Mean & \\
    \cmidrule[.05em](r){1-1} \cmidrule[.05em](lr){2-2} \cmidrule[.05em](lr){3-6} \cmidrule[.05em](lr){7-7} \cmidrule[.05em](lr){8-8} \cmidrule[.05em](lr){9-10} \cmidrule[.05em](lr){11-12} \cmidrule[.05em](lr){13-15} \cmidrule[.05em](l){16-16}
    \cmidrule[.05em](r){1-1} \cmidrule[.05em](lr){2-2} \cmidrule[.05em](lr){3-6} \cmidrule[.05em](lr){7-7} \cmidrule[.05em](lr){8-9} \cmidrule[.05em](lr){10-11} \cmidrule[.05em](lr){12-14} \cmidrule[.05em](l){15-15}   
    Reference & - & - & - & - & - & - & - & - & - & - & - & - & - & $ \mathbf{-2.42} $ & - \\
    \cmidrule(r){1-1} \cmidrule(lr){2-2} \cmidrule(lr){3-6} \cmidrule(lr){7-7} \cmidrule(lr){8-8} \cmidrule(lr){9-10} \cmidrule(lr){11-12} \cmidrule(lr){13-15} \cmidrule(l){16-16}
    Baseline & $ 51.8 $ & $ 19.9 $ & $ 32.4 $ & - & $ 16.9 $ & $ -0.98 $ & $ 0.985 $ & $ 0.812 $ & $ 0.868 $ & $ 0.551 $ & $ 0.677 $ & $ -1.54 $ & $ -4.38 $ & $ -2.62 $ & $ 1020.1 $ \\
    DBS & $ 60.8 $ & $ 19.0 $ & $ 29.6 $ & $ 34.7 $ & $ 15.2 $ & $ 3.12 $ & $ 0.970 $ & $ 0.808 $ & $ 0.862 $ & $ 0.544 $ & $ 0.656 $ & $ -1.51 $ & $ -4.34 $ & $ -2.63 $ & $ 690.8 $ \\
    Nucleus & $ 65.1 $ & $ 18.9 $ & $ 30.8 $ & $ 34.9 $ & $ 15.8 $ & $ 6.28 $ & $ 0.992 $ & $ 0.811 $ & $ 0.860 $ & $ 0.546 $ & $ 0.651 $ & $ -1.56 $ & $ -4.04 $ & $ -2.59 $ & $ 657.1 $ \\
    $k$NN-MT & $ 52.8 $ & $ 21.4 $ & $ 35.1 $ & $ 37.4 $ & $ 17.8 $ & - & $ 0.962 $ & $ 0.817 $ & $ 0.879 $ & $ 0.561 $ & $ 0.711 $ & $ -1.49 $ & $ -4.46 $ & $ -2.62 $ & $ 65.6 $ \\
\bottomrule
  \end{tabular}
  \caption{BSD domain in Japanese-English}\label{tab:bsd}
\end{table*}

\begin{table*}[htbp]
  \centering
  \small
  \tabcolsep 2.3pt
  \begin{tabular}{@{}lcccccccccccccccc@{}}
    \toprule
    & \multirow{2}{*}{DP} & \multicolumn{4}{c}{BLEU} & \multirow{2}{*}{DEQ} & \multirow{2}{*}{MLen} & \multicolumn{2}{c}{COMET} & \multicolumn{2}{c}{BERTScore} & \multicolumn{3}{c}{PLL} & \multirow{2}{*}{Speed}\\
    Method & & @1 & @20 & Mrg & Ref & & & @1 & @20 & @1 & @20 & Max & Min & Mean & \\
    \cmidrule[.05em](r){1-1} \cmidrule[.05em](lr){2-2} \cmidrule[.05em](lr){3-6} \cmidrule[.05em](lr){7-7} \cmidrule[.05em](lr){8-8} \cmidrule[.05em](lr){9-10} \cmidrule[.05em](lr){11-12} \cmidrule[.05em](lr){13-15} \cmidrule[.05em](l){16-16}
    Reference & - & - & - & - & - & - & - & - & - & - & - & - & - & $ -2.93 $ & - \\
    \cmidrule(r){1-1} \cmidrule(lr){2-2} \cmidrule(lr){3-6} \cmidrule(lr){7-7} \cmidrule(lr){8-8} \cmidrule(lr){9-10} \cmidrule(lr){11-12} \cmidrule(lr){13-15} \cmidrule(l){16-16}
    Baseline & $ 32.1 $ & $ 39.6 $ & $ 51.4 $ & - & $ 37.0 $ & $ 0.59 $ & $ 0.981 $ & $ 0.846 $ & $ 0.876 $ & $ 0.714 $ & $ 0.787 $ & $ -2.18 $ & $ -3.53 $ & $ -2.80 $ & $ 1045.7 $ \\
    DBS & $ 41.3 $ & $ 38.9 $ & $ 47.6 $ & $ 53.0 $ & $ 33.9 $ & $ 2.60 $ & $ 0.969 $ & $ 0.844 $ & $ 0.873 $ & $ 0.710 $ & $ 0.772 $ & $ -2.09 $ & $ -3.77 $ & $ -2.84 $ & $ 720.7 $ \\
    Nucleus & $ 55.1 $ & $ 38.1 $ & $ 50.6 $ & $ 54.2 $ & $ 32.5 $ & $ 4.60 $ & $ 0.986 $ & $ 0.847 $ & $ 0.874 $ & $ 0.712 $ & $ 0.782 $ & $ -2.04 $ & $ -3.67 $ & $ -2.80 $ & $ 620.3 $ \\
    $k$NN-MT & $ 31.7 $ & $ 40.6 $ & $ 52.2 $ & $ 52.9 $ & $ 37.6 $ & - & $ 0.986 $ & $ 0.847 $ & $ 0.877 $ & $ 0.719 $ & $ 0.788 $ & $ -2.17 $ & $ -3.51 $ & $ -2.79 $ & $ 8.9 $ \\
    \cmidrule(r){1-1} \cmidrule(lr){2-2} \cmidrule(lr){3-6} \cmidrule(lr){7-7} \cmidrule(lr){8-8} \cmidrule(lr){9-10} \cmidrule(lr){11-12} \cmidrule(lr){13-15} \cmidrule(l){16-16}
    DBS+$k$NN-MT & $ 41.7 $ & $ 39.7 $ & $ 48.7 $ & $ 53.5 $ & $ 34.3 $ & $ 3.10 $ & $ 0.976 $ & $ 0.846 $ & $ 0.873 $ & $ 0.713 $ & $ 0.775 $ & $ -2.03 $ & $ -3.77 $ & $ -2.82 $ & $ 9.0 $ \\
\bottomrule
  \end{tabular}
  \caption{General domain in German-English}\label{tab:general_deen}
\end{table*}

\begin{table*}[htbp]
  \centering
  \small
  \tabcolsep 2.3pt
  \begin{tabular}{@{}lcccccccccccccccc@{}}
    \toprule
    & \multirow{2}{*}{DP} & \multicolumn{4}{c}{BLEU} & \multirow{2}{*}{DEQ} & \multirow{2}{*}{MLen} & \multicolumn{2}{c}{COMET} & \multicolumn{2}{c}{BERTScore} & \multicolumn{3}{c}{PLL} & \multirow{2}{*}{Speed}\\
    Method & & @1 & @20 & Mrg & Ref & & & @1 & @20 & @1 & @20 & Max & Min & Mean & \\
    \cmidrule[.05em](r){1-1} \cmidrule[.05em](lr){2-2} \cmidrule[.05em](lr){3-6} \cmidrule[.05em](lr){7-7} \cmidrule[.05em](lr){8-8} \cmidrule[.05em](lr){9-10} \cmidrule[.05em](lr){11-12} \cmidrule[.05em](lr){13-15} \cmidrule[.05em](l){16-16}
    Reference & - & - & - & - & - & - & - & - & - & - & - & - & - & $ -3.17 $ & - \\
    \cmidrule(r){1-1} \cmidrule(lr){2-2} \cmidrule(lr){3-6} \cmidrule(lr){7-7} \cmidrule(lr){8-8} \cmidrule(lr){9-10} \cmidrule(lr){11-12} \cmidrule(lr){13-15} \cmidrule(l){16-16}
    Baseline & $ 40.7 $ & $ 20.9 $ & $ 30.6 $ & - & $ 19.0 $ & $ -5.75 $ & $ 0.895 $ & $ 0.792 $ & $ 0.846 $ & $ 0.534 $ & $ 0.646 $ & $ -2.01 $ & $ -4.59 $ & $ -3.12 $ & $ 984.3 $ \\
    DBS & $ 59.8 $ & $ 19.6 $ & $ 28.3 $ & $ 32.8 $ & $ 15.6 $ & $ 5.85 $ & $ 0.854 $ & $ 0.786 $ & $ 0.840 $ & $ 0.526 $ & $ 0.630 $ & $ -1.76 $ & $ -4.67 $ & $ -3.07 $ & $ 638.8 $ \\
    Nucleus & $ 71.1 $ & $ 20.4 $ & $ 30.6 $ & $ 33.9 $ & $ 15.5 $ & $ 8.85 $ & $ 0.906 $ & $ 0.785 $ & $ 0.839 $ & $ 0.527 $ & $ 0.634 $ & $ -1.78 $ & $ -4.67 $ & $ -3.07 $ & $ 562.0 $ \\
    $k$NN-MT & $ 40.5 $ & $ 20.8 $ & $ 30.5 $ & $ 31.9 $ & $ 18.9 $ & - & $ 0.898 $ & $ 0.789 $ & $ 0.844 $ & $ 0.529 $ & $ 0.643 $ & $ -1.97 $ & $ -4.57 $ & $ -3.09 $ & $ 8.9 $ \\
    \cmidrule(r){1-1} \cmidrule(lr){2-2} \cmidrule(lr){3-6} \cmidrule(lr){7-7} \cmidrule(lr){8-8} \cmidrule(lr){9-10} \cmidrule(lr){11-12} \cmidrule(lr){13-15} \cmidrule(l){16-16}
    DBS+$k$NN-MT & $ 62.0 $ & $ 19.1 $ & $ 28.2 $ & $ 32.9 $ & $ 15.1 $ & $ 5.57 $ & $ 0.848 $ & $ 0.784 $ & $ 0.840 $ & $ 0.520 $ & $ 0.624 $ & $ -1.62 $ & $ -4.72 $ & $ -3.03 $ & $ 9.2 $ \\
\bottomrule
  \end{tabular}
  \caption{General domain in Japanese-English}\label{tab:general_jaen}
\end{table*}

\begin{table*}[htbp]
  \centering
  \small
  \tabcolsep 2.3pt
  \begin{tabular}{@{}lcccccccccccccccc@{}}
    \toprule
    & \multirow{2}{*}{DP} & \multicolumn{4}{c}{BLEU} & \multirow{2}{*}{DEQ} & \multirow{2}{*}{MLen} & \multicolumn{2}{c}{COMET} & \multicolumn{2}{c}{BERTScore} & \multicolumn{3}{c}{PLL} & \multirow{2}{*}{Speed}\\
    Method & & @1 & @20 & Mrg & Ref & & & @1 & @20 & @1 & @20 & Max & Min & Mean & \\
    \cmidrule[.05em](r){1-1} \cmidrule[.05em](lr){2-2} \cmidrule[.05em](lr){3-6} \cmidrule[.05em](lr){7-7} \cmidrule[.05em](lr){8-8} \cmidrule[.05em](lr){9-10} \cmidrule[.05em](lr){11-12} \cmidrule[.05em](lr){13-15} \cmidrule[.05em](l){16-16}
    Reference & - & - & - & - & - & - & - & - & - & - & - & - &- & $ -2.40 $ & - \\
    \cmidrule(r){1-1} \cmidrule(lr){2-2} \cmidrule(lr){3-6} \cmidrule(lr){7-7} \cmidrule(lr){8-8} \cmidrule(lr){9-10} \cmidrule(lr){11-12} \cmidrule(lr){13-15} \cmidrule(l){16-16}
    Baseline & $ 39.8 $ & $ 30.0 $ & $ 41.7 $ & - & $ 27.0 $ & $ 5.67 $ & $ 0.933 $ & $ 0.865 $ & $ 0.914 $ & $ 0.634 $ & $ 0.739 $ & $ -1.62 $ & $ -3.42 $ & $ -2.40 $ & $ 974.0 $ \\
    DBS & $ 52.8 $ & $ 28.2 $ & $ 37.8 $ & $ 43.6 $ & $ 23.4 $ & $ 3.64 $ & $ 0.916 $ & $ 0.853 $ & $ 0.911 $ & $ 0.624 $ & $ 0.713 $ & $ -1.49 $ & $ -3.68 $ & $ -2.43 $ & $ 755.9 $ \\
    Nucleus & $ 62.3 $ & $ 29.0 $ & $ 40.1 $ & $ 44.4 $ & $ 23.1 $ & $ 5.81 $ & $ 0.941 $ & $ 0.859 $ & $ 0.909 $ & $ 0.629 $ & $ 0.721 $ & $ -1.50 $ & $ -3.64 $ & $ -2.42 $ & $ 669.7 $ \\
    $k$NN-MT & $ 39.7 $ & $ 30.0 $ & $ 41.6 $ & $ 43.0 $ & $ 27.0 $ & - & $ 0.928 $ & $ 0.863 $ & $ 0.914 $ & $ 0.633 $ & $ 0.738 $ & $ -1.61 $ & $ -3.47 $ & $ -2.41 $ & $ 21.0 $ \\
    \cmidrule(r){1-1} \cmidrule(lr){2-2} \cmidrule(lr){3-6} \cmidrule(lr){7-7} \cmidrule(lr){8-8} \cmidrule(lr){9-10} \cmidrule(lr){11-12} \cmidrule(lr){13-15} \cmidrule(l){16-16}
    DBS+$k$NN-MT & $ 54.0 $ & $ 28.5 $ & $ 37.9 $ & $ 43.8 $ & $ 23.2 $ & $ 3.79 $ & $ 0.912 $ & $ 0.853 $ & $ 0.910 $ & $ 0.625 $ & $ 0.713 $ & $ -1.45 $ & $ -3.78 $ & $ -2.44 $ & $ 21.3 $ \\
\bottomrule
  \end{tabular}
  \caption{General Domain in Ukrainian-Czech}\label{tab:general_ukcs}
\end{table*}

\begin{figure*}[t]
\centering
\small
\begin{tabular}{@{}p{7.5cm}p{7.5cm}@{}}
        \toprule[1.5pt]
\mbox{\textbf{Test Input}: \emph{\Ja{コロナに関しまして。}}}          & \\
\mbox{\textbf{Reference}: \emph{I have a question about COVID.}} & \\
                     \midrule[0.75pt]
 \textbf{DBS+$k$NN-MT+Randomize} & \textbf{DBS} \\ 
                     \midrule[0.75pt]
\emph{About corona.} & \emph{Regarding corona.} \\ \addlinespace[0.1em]
\emph{With regards to corona.} & \emph{About corona.} \\ \addlinespace[0.1em]
\emph{About \underline{COVID-19}.} & \emph{It is about corona.} \\ \addlinespace[0.1em]
\emph{Regarding corona.} & \emph{We are talking about corona.} \\ \addlinespace[0.1em]
\emph{With regards to \underline{COVID-19}.} & \emph{With regards to corona.} \\
\end{tabular}
    \caption{Full example 20-best lists using DBS-based methods}
    \label{fig:ex_full}
\end{figure*}

\end{document}